\documentclass[letterpaper, 10 pt, conference]{ieeeconf}  

\IEEEoverridecommandlockouts                              

\usepackage{times}
\usepackage{epsfig}
\usepackage{graphicx}
\usepackage{amsmath}
\usepackage{amssymb}
\usepackage{pdfpages}
\usepackage[font={small}]{subcaption}
\usepackage[font={small}]{caption}
\usepackage{multirow}
\usepackage{tabu}
\usepackage{cite}
\usepackage[breaklinks=true,colorlinks,bookmarks=false]{hyperref}

\newcommand\Tstrut{\rule{0pt}{2.1ex}}         
\newcommand\Bstrut{\rule[-0.5ex]{0pt}{0pt}}   

\title{\LARGE \bf
DirectShape: Direct Photometric Alignment of Shape Priors \\for Visual Vehicle Pose and Shape 
Estimation}

\author{Rui~Wang$^{1}$, Nan~Yang$^{1}$, J\"org~St\"uckler$^{2}$, Daniel~Cremers$^{1}$
\thanks{$^{1}$R. Wang, N. Yang and D. Cremers are with the Department of Computer Science, 
Technical University of Munich, Garching bei M\"unchen, 85748, Germany and Artisense Corporation, 
350 Cambridge Avenue 250, Palo Alto, CA 94306, USA. {\tt\small \{wangr, yangn,  
cremers\}@in.tum.de}}%
\thanks{$^{2}$J. St\"uckler is with Max Planck Institute for Intelligent Systems T\"ubingen, 
T\"ubingen, 72076, Germany.\protect\\{\tt\small joerg.stueckler@tuebingen.mpg.de}}%
}

\begin{document}

\maketitle
\thispagestyle{empty}
\pagestyle{empty}

\begin{abstract}	
Scene understanding from images is a challenging problem encountered in autonomous 
driving. On the object level, while 2D methods have gradually evolved from computing simple 
bounding boxes to delivering finer grained results like instance segmentations, the 3D family is 
still dominated by estimating 3D bounding boxes. In this paper, we propose a novel approach to 
jointly infer the 3D rigid-body poses and shapes of vehicles from a stereo image pair using shape 
priors. Unlike previous works that geometrically align shapes to point clouds from dense 
stereo reconstruction, our approach works directly on images by combining a photometric and a 
silhouette alignment term in the energy function. An adaptive sparse point selection scheme is 
proposed to efficiently measure the consistency with both terms. In experiments, we show superior 
performance of our method on 3D pose and shape estimation over the previous geometric approach 
and demonstrate that our method can also be applied as a refinement step and significantly 
boost the performances of several state-of-the-art deep learning based 3D object detectors. All 
related materials and demonstration videos are available at the project page
\url{https://vision.in.tum.de/research/vslam/direct-shape}.
\end{abstract}

\section{INTRODUCTION}
3D scene understanding is a fundamental task with widespread applications in robotics, augmented 
reality and autonomous driving. For autonomous vehicles it is critical to observe the poses and 3D 
shapes of other cars for navigation planning and control. Yet the inference of such object 
properties from images is a challenging task due to camera projection, variability in view-point, 
appearance and lighting condition, transparent or reflective non-lambertian surfaces on cars, etc. 
The community therefore has so far mainly cogitated upon estimating bounding boxes which only 
contain coarse information on the object poses and sizes. Although with the advances in computer 
vision 2D object detection has gradually evolved to delivering finer grained results such as 
instance segmentations, 3D methods are still focusing on estimated bounding boxes. 

\begin{figure}[t]
	\centering
	\includegraphics[width=0.47\textwidth]{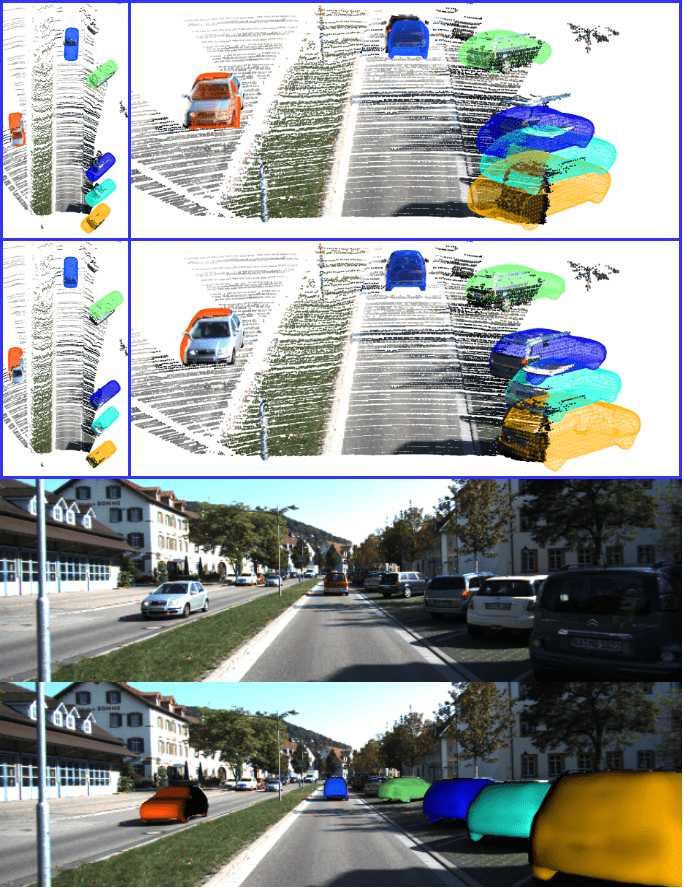}
	\caption{We propose to jointly estimate 3D vehicle poses and shapes directly on image 
	intensities using shape priors. Top: The initial and our estimated 3D poses and shapes in 3D. Note 
	that the LiDAR point clouds are only for visualization and are not used in our 
	method. Bottom: The input image and its overlay with our results.}
	\label{fig:teaser}
	\vspace{-4.5ex}
\end{figure}

In this paper, we address joint 3D rigid-body pose estimation and shape reconstruction from a 
single stereo image pair using 3D shape priors, as illustrated in Fig.~\ref{fig:teaser}. Shape priors allow 
for confining the search space 
over possible shapes and make vision based pose and shape estimation more robust to effects such as 
occlusion, lighting conditions or reflective surfaces. We use volumetric signed distance functions 
(SDF) to implicitly represent the shape of exemplar car models and a linear low-dimensional 
subspace embedding of the shapes. While previous works made it possible to align the 3D shape 
geometrically to point clouds estimated by stereo reconstruction 
\cite{engelmann2016joint,engelmann2017samp}, we infer shapes and poses directly from images, thus 
avoid introducing the errors from stereo matching into the pipeline. In our case, the shape prior 
is aligned with detected cars in stereo images using photometric and silhouette consistency 
constraints which we formulate as a non-linear least squares problem and optimize using the 
Gauss-Newton method. Experiments demonstrate superior performance of our method over the previous 
approach that uses geometric alignment with dense stereo reconstructions. Moreover, as learning 
based 3D object detectors have become more popular, we also show that our method can be 
applied as a refinement step that significantly boosts the performances of all the tested methods. 

In summary, our contributions are:
\begin{itemize}
	\itemsep0em 
	\item A novel approach for joint 3D pose and shape estimation that delivers more precise and 
	fine grained results than 3D bounding boxes that are commonly estimated by most of the current 
	methods. 
	\item  Our approach works directly in image space and thus avoids introducing errors from 
	intermediate steps. It delivers superior performance over the previous approach that uses a 
	geometric formulation for alignment.
	\item A fully differentiable formulation that operates directly between image and SDF 
	based 3D shape embedding.
	\item Our method can be applied together with state-of-the-art learning based approaches and 
	significantly boosts the performances of all the tested methods. 
\end{itemize}

\section{Related Work}
\begin{figure*}[t]
	\centering
	\includegraphics[width=0.98\textwidth]{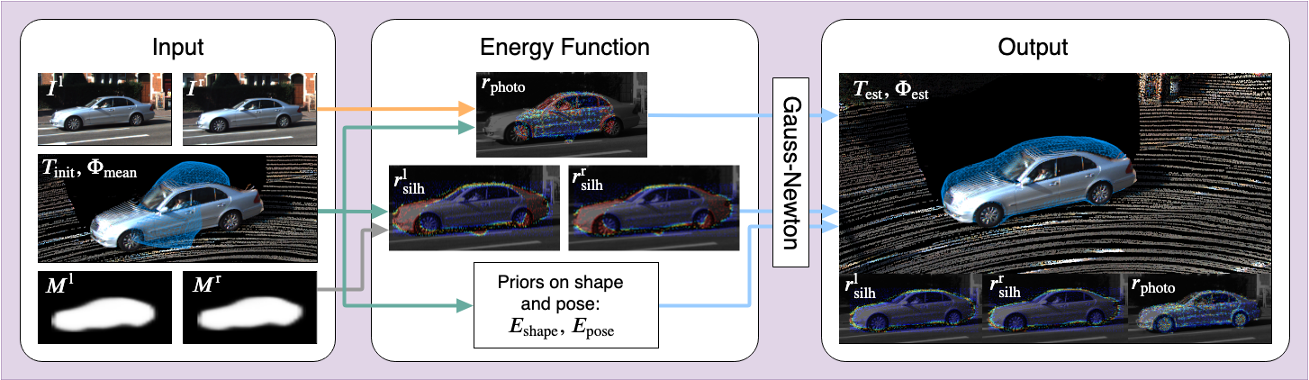}
	\caption{System overview. As input our method takes a stereo frame $\mathrm{I}^{l}$, 
	$\mathrm{I}^{r}$, an initial object pose $\mathrm{T}_{init}$, the learned mean shape 
	$\mathrm{\Phi}_{mean}$, and the object segmentation masks $\mathrm{M}^{l}$, $\mathrm{M}^{r}$. 
	Based on the current pose and shape, the object is projected to $\mathrm{I}^{l}$ and 
	$\mathrm{I}^{r}$ and the consistencies between the projections and the segmentation masks are 
	measured by the silhouette alignment residuals $\mathrm{r}_{silh}$ 
	(Sec.\ref{subsec:segmentaton_term}). Meanwhile, the object pixels in $\mathrm{I}^{l}$ can be 
	warped to $\mathrm{I}^{r}$. The color 
	consistencies are measured by the photometric consistency residuals $\mathrm{r}_{photo}$ 
	(Sec.\ref{subsec:photo_term}). The two terms together with the prior terms (Sec.\ref{subsec:prior}) 
	are formulated as a non-linear energy function and optimized using the 
	Gauss-Newton method. As output, our method delivers refined object pose $\mathrm{T}_{est}$ and 
	shape $\mathrm{\Phi}_{est}$.}
	\vspace{-3ex}
	\label{fig:system_overview}
\end{figure*}

{\bf 3D object detection.} Many successful object detectors have been focusing on localizing 
objects by 2D bounding boxes~\cite{felzenszwalb2009object, ren2015_fasterrcnn, dai2016r, 
dai2017deformable, redmon2017_yolo9000} and later by segmentation masks~\cite{chen2017deeplab, 
he2017mask, kirillov2019panoptic} in images. As object detection in 2D matures, the community 
starts to target at the much more challenging 3D object detection task~\cite{satkin2013_3dnn, 
3dopNIPS15, shuran2014_3dslidingshapes, cvpr16chen, mousavian20173d, xu2018multi, licvpr2019, 
wang2019pseudo, kundu20183d}. 
Chen et al.~\cite{3dopNIPS15} present 3DOP 
that first generates 3D object proposals in stereo reconstructions and then scores them in 
the 2D image using Faster-RCNN~\cite{ren2015_fasterrcnn}. By combining the 3D orientations and 
dimensions regressed by a network with 2D geometric constraints, Mousavian et 
al.~\cite{mousavian20173d} largely improve the stability and accuracy of 3D detection. While 
recent deep learning based methods continue to boost the quality of the estimated 3D bounding 
boxes~\cite{xu2018multi, licvpr2019, wang2019pseudo}, Kundu et al.~\cite{kundu20183d} first propose 
to use a network to regress the 3D pose and shape at the same time, yet the shape is not 
evaluated in their paper. At the current stage, learning based approaches still can only provide a 
coarse estimate of the object pose and shape. Moreover, it is an open research question how to 
assess the quality of learning based detections. Optimization based methods can refine coarse 
detections and introspect the quality of the fit of the measurements to the model. We believe that 
is where they come to the stage. Based on the coarse 3D poses estimated by the learning based 
approaches, our optimization pipeline jointly refines the poses and estimates precise 3D shapes, 
which contain much more information than 3D bounding boxes and we believe are more useful for 
applications such as obstacle avoidance and new view synthesis.

{\bf 3D scene understanding.} The availability of large-scale 3D model 
databases, capable 3D object detectors and fast rendering techniques have 
spawned novel interest in the use of geometric methods for 3D scene 
understanding.
Salas-Moreno et al.~\cite{salasmoreno2013_slampp} integrate object instances into RGB-D SLAM. The 
object instances are included as additional 
nodes in pose graph optimization which finds a consistent camera trajectory and object pose 
estimate. Geiger and Wang~\cite{Geiger2015GCPR} infer 3D object and scene layout from a single 
RGB-D image by aligning CAD object models with the scene. The approach in~\cite{Ortizcayon3DV16} 
detects cars using a CNN-based detector, estimates dense depth using multi-view stereo and aligns a 
3D CAD model to the detected car using depth and silhouette constraints.
Closely related to our approach, Engelmann et al.~\cite{engelmann2016joint,engelmann2017samp} use 
3D shape embeddings to determine pose and shape of cars which are initially detected by 
3DOP~\cite{3dopNIPS15}. They also embed volumetric signed distance fields (SDF) of CAD models 
using PCA and formulate a non-linear least squares problem.
Their data term, however, relies on a dense stereo reconstruction and measures the distance of 
reconstructed points to the object surface. It is thus susceptible to the errors of 
the black-box stereo reconstruction algorithm. Our approach does not require dense stereo 
matching but directly fits 3D SDF shape embeddings through photometric and silhouette 
alignment to the stereo images. While silhouette alignment has been used previously to align 3D 
object models~\cite{Sandhu2011_kernelsegposerecon, Prisacariu2012_segposerecon, DameCVPR13}, these 
methods mainly focus on controlled settings such as only one dominated object appears in the image. 
By combining silhouette alignment with photometric alignment and explicitly addressing occlusions, 
we target at the much more challenging real-world traffic scenarios.

\section{Proposed Method}
\subsection{Notations}
Throughout this paper, $\mathbf{p}$ and $\mathbf{X}$ respectively denote image pixels 
and 3D points. Subscripts $o$ and $c$ define coordinates in object and camera coordinate system. 3D 
rigid body transformations $\mathbf{T}^{b}_{a} = [\mathbf{R}^{b}_{a}, \mathbf{t}^{b}_{a}; 0, 1] \in 
\text{SE}(3)$ transforms coordinates from system $a$ to system $b$, where $\mathbf{R}$ and 
$\mathbf{t}$ are the 3D rotation matrix and translation vector. In our optimization, 3D poses are 
represented by their twist coordinates in Lie-algebra $\boldsymbol{\xi} \in \mathfrak{se}(3)$ and 
3D shapes are represented by a SDF voxel grid $\mathbf{\Phi}$. PCA is performed to embed 3D shapes 
into a low dimensional space~\cite{engelmann2016joint,engelmann2017samp} and thus each shape can be 
represented by $\mathbf{\Phi(\mathbf{z})} = \mathbf{Vz} + \mathbf{\Phi}_{mean}$, where  
$\mathbf{V}$ is the transpose of the subspace projection matrix, $\mathbf{z}$ the shape encoding 
vector and $\mathbf{\Phi}_{mean}$ the mean shape of the gathered object set. 
The SDF value of a location within the 3D grid is obtained by trilinear interpolation to achieve subvoxel 
accuracy.
While more sophisticated nonlinear shape encodings like Kernel PCA and  
GP-LVM~\cite{dambreville2008framework, sandhu2011nonrigid,prisacariu2011nonlinear, 
prisacariu2011shared, prisacariu2012simultaneous, zheng2015object} exist, we find for cars the PCA 
model is sufficient. 
Nevertheless, our formulation is agnostic to the shape 
representation, thus can be easily adopted for other SDF based shape encodings.    

\subsection{System Overview}
An overview of our system is shown in Fig.~\ref{fig:system_overview}. Given a stereo frame with the 
object segmentations in both images, the quality of the current estimate of the object pose and 
shape can be measured by two energy terms: (1) by projecting the current shape to both images, a 
silhouette alignment term ${E}_{silh}$ measures the consistencies of the projections with the 
corresponding segmentation masks; (2) object pixels in the left image are warped to the right 
image and the photometric consistency term ${E}_{photo}$ measures the color differences. Based on 
domain knowledge, we add priors terms on the object pose and shape. Our final energy function 
combines the terms above and is optimized using the Gauss-Newton method. In the following we 
present the details of each energy term.

\subsection{Silhouette Alignment Term}
\label{subsec:segmentaton_term}

The silhouette alignment term measures the consistency between the image segmentation masks
$\mathbf{M}^{l/r}$ and the object masks obtained by projecting the 3D SDF shape embedding 
$\mathbf{\Phi}$ into the images based on its current shape and pose estimate. Denoting the value of 
the shape projection mask at pixel $\mathbf{p}$ by $\pi(\mathbf{\Phi},\mathbf{p})$ (details later) 
which holds values close to 1 inside and 0 outside the object, the consistency with 
$\mathbf{M}^{l/r}$ can be expressed by
\begin{equation}
E_{silh}^{l/r} = \frac{1}{|\Omega|}\sum_{\mathbf{p} \in \Omega} r_{silh}^{l/r}(\mathbf{p}),
\end{equation}
\begin{equation}
r_{silh}^{l/r}(\mathbf{p}) = -\text{log} 
\big( \pi(\mathbf{\Phi}, 
\mathbf{p}) 
p_{fg}(\mathbf{p}) + (1 - \pi(\mathbf{\Phi}, \mathbf{p})) p_{bg}(\mathbf{p}) 
\big),
\end{equation}
where $\Omega$ is the set of the pixels of this object instance, $p_{fg}$ and $p_{bg}$ are the 
foreground and background probabilities from $\mathbf{M}^{l/r}$. Ideally, if at $\mathbf{p}$ the 
shape projection coincides with the object segmentation, the value inside the $log$ is close to 1, 
leading to a small silhouette alignment residual $r_{silh}(\mathbf{p})$; Otherwise the value inside 
the $log$ is a positive number close to 0, resulting in a large $r_{silh}(\mathbf{p})$. Examples of 
the silhouette alignment residuals in the left and right images are shown in the middle of 
Fig.~\ref{fig:system_overview} (higher residuals are denoted in red).

Our requirement on the shape projection function $\pi(\mathbf{\Phi},\mathbf{p})$ is its 
differentiability wrt. $\mathbf{\Phi}$ and $\mathbf{p}$. Inspired 
by~\cite{prisacariu2012simultaneous, DameCVPR13}, we define it as
\begin{equation}
\pi(\mathbf{\Phi}, \mathbf{p}) = 1 - \prod_{\mathbf{X}_{o}} 
\frac{1}{e^{\mathbf{\Phi}(\mathbf{X}_{o})\zeta} + 1},
\end{equation}
where $\mathbf{X}_{o}$ are sampled 3D points along the ray through the camera center and 
$\mathbf{p}$, 
$\zeta$ is a constant that defines the smoothness of the projection contour. 

\begin{figure}[t]
	\centering
	\captionsetup[subfigure]{justification=centering}
	\begin{subfigure}[t]{0.155\textwidth}
		\includegraphics[width=\textwidth]{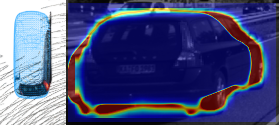}
		\caption{Iteration 1.}
	\end{subfigure}
	\begin{subfigure}[t]{0.155\textwidth}
		\includegraphics[width=\textwidth]{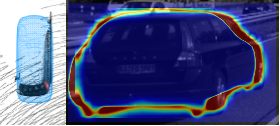}
		\caption{Iteration 2.}
	\end{subfigure}
	\begin{subfigure}[t]{0.155\textwidth}
		\includegraphics[width=\textwidth]{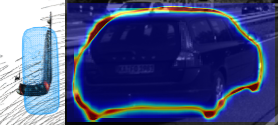}
		\caption{Iteration 3.}
	\end{subfigure}
	\vspace{-1ex}
	\caption{Ambiguity between 3D shape and pose when using only the silhouette alignment term. 
	While fitting the shape projection (harder contour) to the object segmentation (softer 
	contour), the 3D pose drifts away as shown in the bird's-eye view on the left.}\vspace{-4ex}
	\label{fig:projection_drift}
\end{figure}

It is worth noting that the silhouette alignment term only confines the projections of 3D shapes to 
2D silhouettes, which is not sufficient to resolve the ambiguity between 3D shapes and poses. 
Especially in our single-frame stereo setting, even with class-specific priors regularizing the 
estimated pose, the 3D model often drifts away to better fit the 2D 
silhouettes, as illustrated in Fig.~\ref{fig:projection_drift}. We thus propose in the next section 
to further enforce photometric consistency to favor poses and shapes that give less color 
inconsistencies between the left and right images.

\subsection{Photometric Consistency Term}
\label{subsec:photo_term}

By warping the object pixels from the left image $\mathrm{I}^{l}$ to the right $\mathrm{I}^{r}$, 
the photometric consistency term measures the color consistencies. For each pixel within the shape 
projection, we determine the depth to the object surface through raycasting and finding the 
intersection with the zero-level set in the SDF. Using this depth and the current object pose, the 
pixels are transformed from $\mathrm{I}^{l}$ to $\mathrm{I}^{r}$. Under the brightness constancy 
assumption, when the pose and shape estimates of an object are correct, the corresponding pixel 
intensities in the two images should be the same. Our photometric consistency term is formally
defined as:
\begin{equation}
E_{photo} = \frac{1}{|\Omega'| |N_{\mathbf{p}}|} \sum_{\mathbf{p} \in \Omega'} 
\sum_{\tilde{\mathbf{p}} \in N_{\mathbf{p}}} 
\omega_{\mathbf{p}} 
|| r_{photo}(\tilde{\mathbf{p}}) ||_{\gamma},
\end{equation}
\begin{equation}
r_{photo}(\tilde{\mathbf{p}}) = \mathbf{I}_{r}\big(\Pi_{c}(\mathbf{R}_{l}^{r} 
\Pi_{c}^{-1}(\tilde{\mathbf{p}}, 
d_{\mathbf{p}}) + \mathbf{t}_{l}^{r})\big) - 
\mathbf{I}_{l}\big(\tilde{\mathbf{p}}\big),
\end{equation}
where $\Omega'$ is the set of the pixels that have intersecting rays 
with the current shape surface, $N_{\mathbf{p}}$ is a small image neighborhood 
around $\mathbf{p}$. For each pixel $\tilde{\mathbf{p}}$ in $\mathrm{I}^{l}$, we warp it to 
$\mathrm{I}^{r}$ based on the current depth of 
its central pixel $d_\mathbf{p}$ and the relative 3D rotation $\mathbf{R}_{l}^{r}$ 
and translation $\mathbf{t}_{l}^{r}$. 
$\Pi_{c}(\cdot)$ and $\Pi_{c}^{-1}(\cdot)$ are the camera projection and 
back-projection functions that transform 3D coordinates to pixel coordinates 
and vice versa. The photometric residual 
$r_{photo}(\tilde{\mathbf{p}})$ is guarded by the Huber norm 
$||\cdot||_{\gamma}$ and an image gradient based weighting 
$\omega_{\mathbf{p}} = c^{2}/(c^{2} + ||\nabla \mathbf{I}_{l}(\mathbf{p})||_{2}^{2})$, 
where $c$ is a constant and $\nabla \mathbf{I}_{l}(\mathbf{p})$ is the image gradient at 
$\mathbf{p}$. An example of the photometric consistency residuals are shown 
in the middle of Fig.~\ref{fig:system_overview}. 

The idea behind the photometric consistency term is analogous to direct image alignment applied 
in recent direct visual odometry (VO) and SLAM methods~\cite{engel2013semi, engel2014lsd, 
wang2017stereo, engel2018direct}. The difference is that instead of directly optimizing for the 
depth of each pixel independently, in our case the pixel depths are implicitly parameterized by the 
object pose and shape, i.e., $d_{\mathbf{p}} = d(\mathbf{p}, \mathbf{z}, \mathbf{T}^{o}_{c})$, which 
brings challenges when deriving the derivative of $r_{photo}$ wrt. $\mathbf{z}$ and 
$\mathbf{T}^{o}_{c}$. 
Nevertheless, the analytical Jacobians are still achievable and the thorough derivations are provided on 
our project page.

\subsection{Prior Terms} 
\label{subsec:prior}
As cars can only locate on road surface and rotate along the axis that is perpendicular to the road, we 
encode this domain knowledge with two priors for the pose estimation. Besides, since cars 
cannot have randomly diverse shapes, we further regularize the estimated shape to be close to our 
mean shape. Our prior term is therefore defined as:
\begin{equation}
E_{prior} = \lambda_{1} E_{shape} + \lambda_{2} E_{trans} + \lambda_{3} 
E_{rot},
\end{equation}
\begin{equation}
E_{shape} = \sum_{i=1}^{K}(\frac{z_{i}}{\sigma_{i}})^2,
\end{equation}
\begin{equation}
E_{trans} = (\mathbf{t}_{o}^{c}(y) - g(\mathbf{t}_{o}^{c}(x, z))(y))^2,
\end{equation}
\begin{equation}
E_{rot} = (1 - (\mathbf{R}_{o}^{c}[0, -1, 0]^{\top})^{\top} \mathbf{n}_{g})^{2},
\end{equation} 
where $\lambda_{1, 2, 3}$ are scalar weighting factors, $\sigma_{i}$ is 
the Eigenvalue of the i-th principal component;
$g(\mathbf{t}_{o}^{c}(x, z))(y)$ is the height of the road plane
at position $\mathbf{t}_{o}^{c}(x, z)$ so that $E_{trans}$ pulls 
the bottom of the car close to the ground plane; $\mathbf{R}_{o}^{c}[0, -1, 
0]^{\top}$ is the direction vector of the negative object y-axis and $\mathbf{n}_{g}$ is the normal 
vector of the ground surface. 
$E_{rot}$ penalizes a large difference between the two directions. 

\subsection{Adaptive Point Sampling}
\label{sugsection:point_sampling}
\begin{figure}[t]
	\centering
	\includegraphics[width=0.23\textwidth]{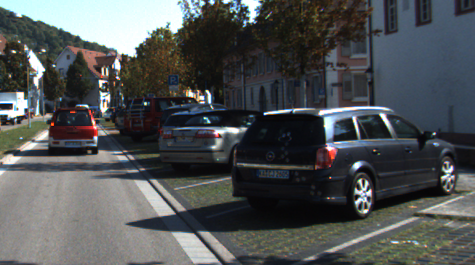}
	\includegraphics[width=0.23\textwidth]{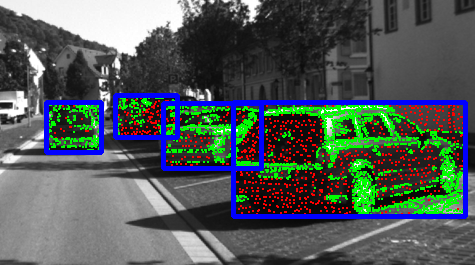}
	\vspace{-0.5ex}
	\caption{Adaptive point sampling. Pixels are sampled to meet the 
		desired density for each object, preferring pixels with high image gradient (green) but 
		meanwhile maintaining a close to uniform distribution (red).}
	\vspace{-3ex}
	\label{fig:sampling}
\end{figure}

In previous works, the silhouette alignment term was computed densely for all the pixels on 
GPUs~\cite{DameCVPR13}. While the same implementation principal can be 
applied to the photometric consistency term, we observe that the dense pixel field contains highly 
redundant information that contributes only minor to both terms. Recent direct VO 
methods adopt the idea to sample pixels with sufficient gradients and meanwhile favor a more 
spatially uniform distribution~\cite{engel2018direct, wang2017stereo, yang2018deep, gao2018ldso}. 
Besides reducing the computational burden, this strategy also suppresses the ambiguous information 
being added to the system. We observe that due to the reflections on the car 
surfaces, this strategy becomes even more relevant in our case and can drastically improve the 
convergence of the photometric term. 
One issue, though, is the sampling strategy proposed in~\cite{engel2018direct} is adaptive and 
sometimes can still give very imbalanced spatial distributions. This is undesired for the 
silhouette alignment if too few pixels are sampled from the object boundary area, as the 
corresponding 3D parts will not be well constrained. We thus modify the adaptive sampling 
in~\cite{engel2018direct} to a two-round pipeline: The image is first discretized into a regular 
grid and a threshold for each cell is computed based on the gradient magnitudes of the pixels 
within it. The image is then re-discretized using smaller cell size and pixels with gradients above 
the threshold are selected (green in  Fig.~\ref{fig:sampling}). This is identical as  
in~\cite{engel2018direct}. 
In the second round, we select the pixel with the highest gradient (red in Fig.~\ref{fig:sampling}) 
for each cell that doesn't get any sample from the previous round. To ensure a consistent density 
for object instances with different sizes in the image, we compute the numbers to sample 
proportionally 
to the area of their bounding boxes as $0.05\times height\times width$.  

\begin{figure*}[t]
	\captionsetup[subfigure]{justification=centering}
	\centering
	\begin{subfigure}[t]{.18\textwidth}
		\includegraphics[width=\linewidth]{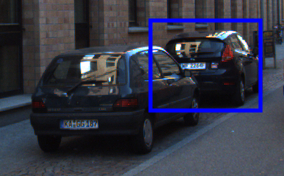}
		\caption{Input image with bounding box.}
		\label{box}
	\end{subfigure}
	\begin{subfigure}[t]{.18\textwidth}
		\includegraphics[width=\linewidth]{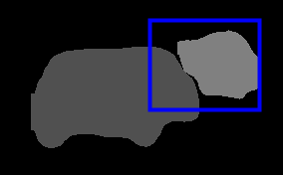}
		\caption{Segmentation masks and bounding box.}
		\label{box_and_seg}
	\end{subfigure}
	\begin{subfigure}[t]{.18\textwidth}
		\includegraphics[width=\linewidth]{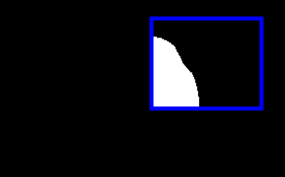}
		\caption{Occlusion mask.}
		\label{occ_mask}
	\end{subfigure}
	\begin{subfigure}[t]{.18\textwidth}
		\includegraphics[width=\linewidth]{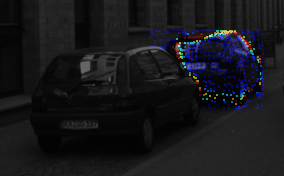}
		\caption{Silhouette alignment residuals.}
		\label{silh_term}		
	\end{subfigure}
	\begin{subfigure}[t]{.18\textwidth}
		\includegraphics[width=\linewidth]{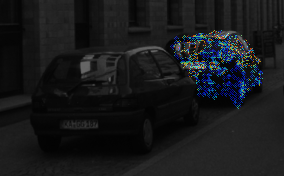}
		\caption{Photometric consistency residuals.}
		\label{photo_term}
	\end{subfigure}
	\vspace{-1ex}
	\caption{Occlusion handling. For each object detection, we check if its 2D 
		bounding box is overlapped by the segmentation mask of any other object that is 
		closer to the camera (\ref{box_and_seg}). Such overlapping part is considered as the 
		occlusion mask (\ref{occ_mask}) and is used in the computation of both the silhouette 
		alignment (\ref{silh_term}) and the photometric consistency residuals 
		(\ref{photo_term}) to exclude pixels from the occluded part.}
	\vspace{-3ex}
	\label{fig:occ}
\end{figure*}

\subsection{Occlusion Handling}
When a car is occluded by other cars (the most common case in traffic scenarios), we can extract an 
occlusion mask using the bounding box of the occluded car and the segmentation masks of the 
occluding cars, as illustrated in Fig~\ref{box}-\ref{occ_mask}. To this end, we sort all 
the cars appearing in the image based on the bottom coordinates of their 2D bounding boxes and thus 
they are ordered roughly according to their distances to the camera. Then for each car we check if 
there is any closer car overlapping with it and if yes we extract the occlusion mask according to 
Fig~\ref{box}-\ref{occ_mask}. The occlusion mask is used in the computations of both the silhouette 
alignment term and the photometric consistency term to exclude the samples from the occluded area, 
as shown in Fig~\ref{silh_term} and~\ref{photo_term}.

\subsection{Optimization}
\label{subsec:optimziation}

Our final energy function is defined as the weighted sum of the previously defined terms
\begin{equation}
E = \lambda_{silh} E_{silh}^{l} + \lambda_{silh} E_{silh}^{r} + E_{photo} + E_{prior},
\end{equation}
where $\lambda_{silh}$ is a scalar weighting factor. Note that all the energy terms 
have quadratic forms except for $E_{silh}$, which prevents the application of 2nd-order  
optimization methods. While in the previous works $E_{silh}$ is typically optimized using 
1st-order methods like gradient descent, we reformulate it as an iteratively reweighted least 
squares problem as
\begin{equation}
E_{silh}^{l/r} = \frac{1}{|\Omega|}\sum_{\mathbf{p} \in \Omega} 
\omega_{\mathbf{p}}'(r_{silh}^{l/r}(\mathbf{p}))^{2},
\end{equation}
where $\omega_{\mathbf{p}}' = 1/r_{silh}{(\mathbf{p})}$ is recalculated in each iteration based on 
the value of $r_{silh}(\mathbf{p})$ of the current iteration. $E$ is thus optimized using 
Gauss-Newton for the variables $[\boldsymbol{\xi}_{c}^{o}; \mathbf{z}]$, where 
$\boldsymbol{\xi}_{c}^{o}$ are the twist coordinates of the 3D rigid-body pose of the object in 
the camera coordinate system and $\mathbf{z}$ is the shape encoding vector. 
It is worth pointing out that previous works~\cite{zhu2018object, kundu20183d} stated that the 
process of rendering from 3D shape to image is not differentiable due to the non-linearity and 
hidden relationship between the two domains, and thus opt for workarounds such as finite 
difference. We claim that such process is actually fully differentiable. Please refer to our project page 
for the details of all the analytical Jacobians.

\section{Experiments}
Our method aims at estimating more precise and accurate geometric properties than 3D bounding 
boxes for cars. To demonstrate this ability, we separately evaluate its performances on 3D shape 
estimation and 3D pose refinement, where the KITTI Stereo 2015~\cite{Menze2015CVPR} and 3D 
Object~\cite{Geiger2012CVPR} benchmarks are adopted for the two tasks respectively. Throughout all 
our experiments, object masks are generated by Mask-RCNN~\cite{he2017mask}. We set 
$\lambda_{silh}=12$, $\lambda_{1}=10$, $\lambda_{2}=10$ and $\lambda_{3}=10^7$ and 
$\mathrm{\Phi}_{mean}$ is used as shape initialization. The 
CPU implementation of our optimization runs at around 100ms per object and 180ms per frame on 
KITTI. Porting to GPU may yield further runtime improvement.  

\subsection{Shape Estimation}
\begin{figure}
	\centering
	\includegraphics[width=0.40\textwidth]{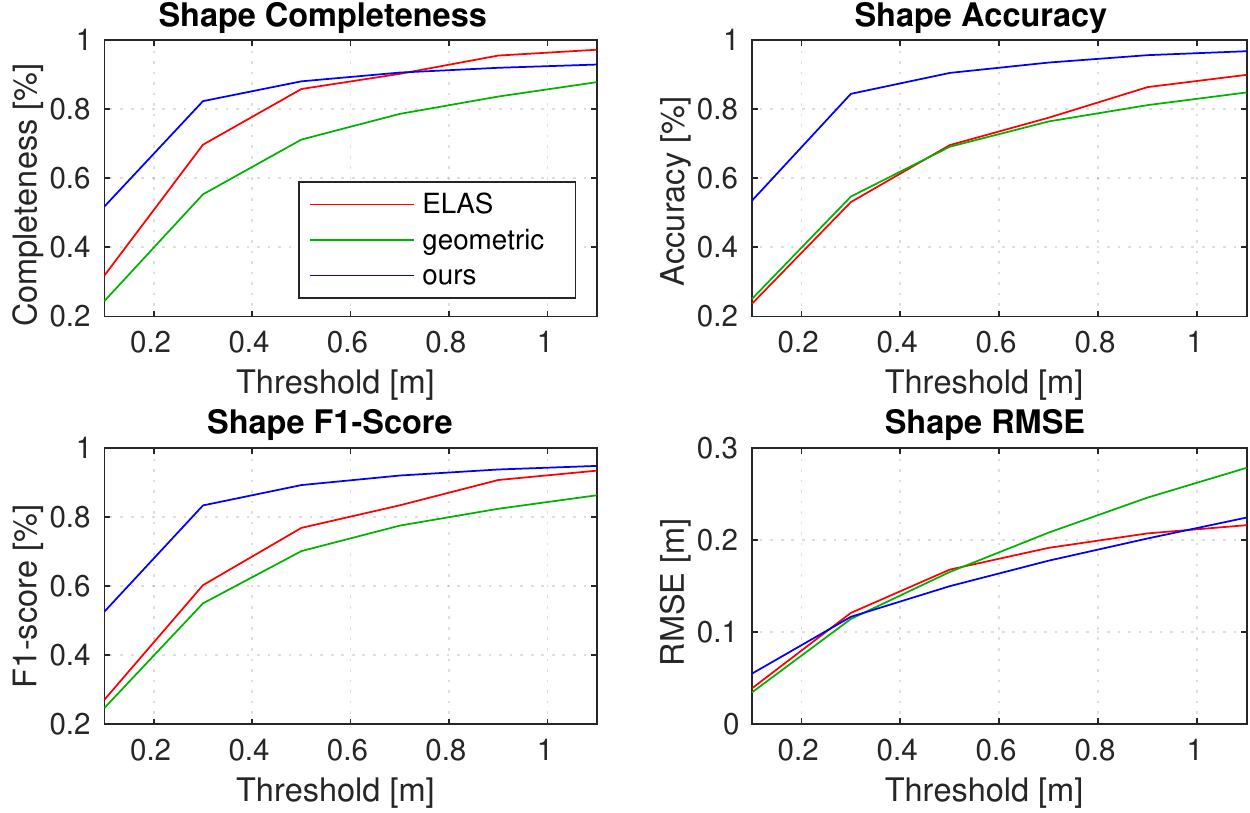}
	\vspace{-1ex}
	\caption{Quantitative evaluation on shape reconstruction. Our approach 
		outperforms the geometric approach~\cite{engelmann2016joint} 
		in all measures.}
	\label{fig:eval_shape}
	\vspace{-2.0ex}
\end{figure}

\begin{figure}[]
	\centering
	\includegraphics[width=0.40\textwidth]{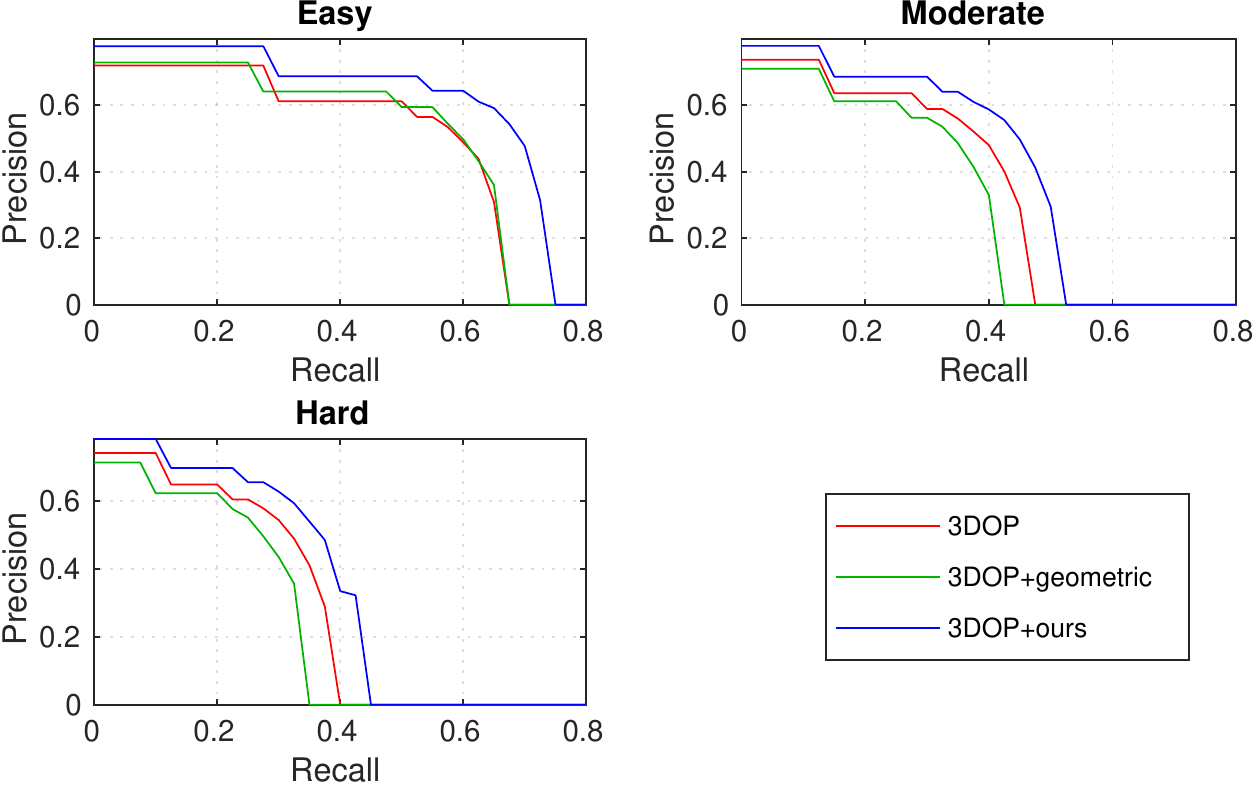}
	\vspace{-1ex}
	\caption{Quantitative evaluation on pose estimation in comparison to 
		the geometric approach~\cite{engelmann2016joint}. IoU=0.5 is used for computing these 
		precision-recall curves.}\vspace{-2.5ex}
	\label{fig:eval_pose}
\end{figure}

\begin{figure}[!]
	\captionsetup[subfigure]{justification=centering,labelformat=empty}
	\centering
	\includegraphics[width=0.091\textwidth]{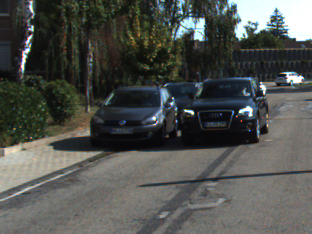}
	\includegraphics[width=0.091\textwidth]{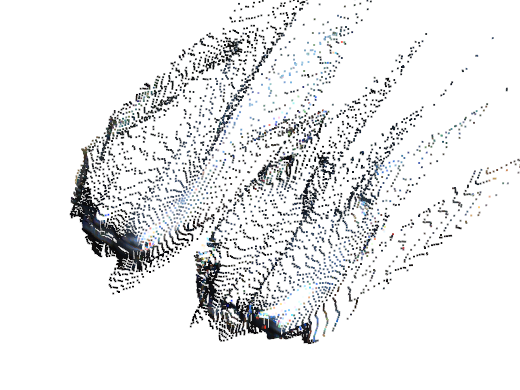}
	\includegraphics[width=0.091\textwidth]{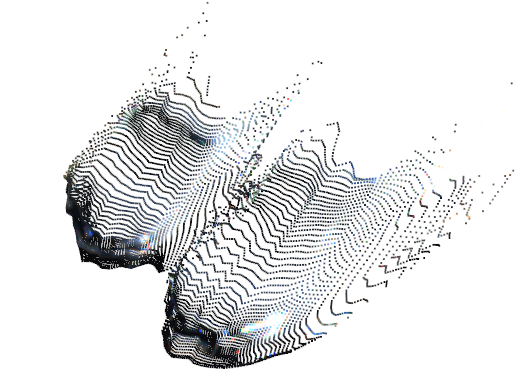}
	\includegraphics[width=0.091\textwidth]{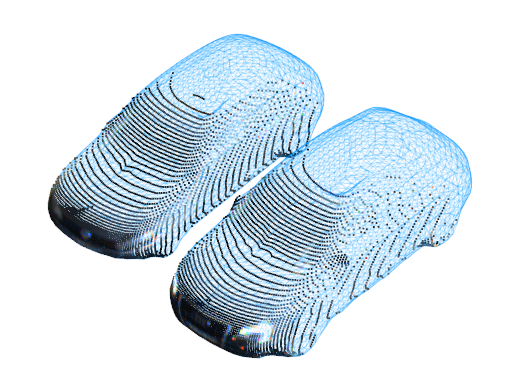}
	\includegraphics[width=0.091\textwidth]{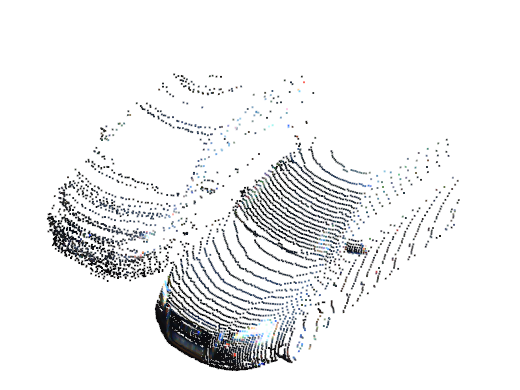}
	
	\includegraphics[width=0.091\textwidth]{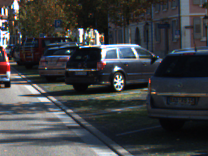}
	\includegraphics[width=0.091\textwidth]{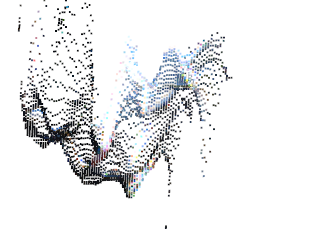}
	\includegraphics[width=0.091\textwidth]{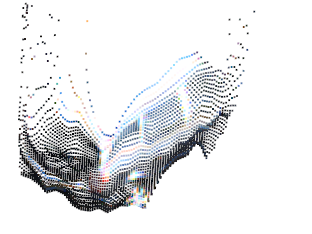}
	\includegraphics[width=0.091\textwidth]{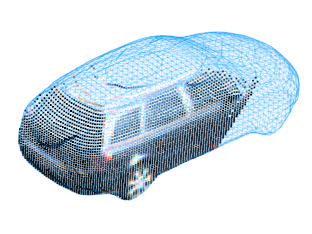}
	\includegraphics[width=0.091\textwidth]{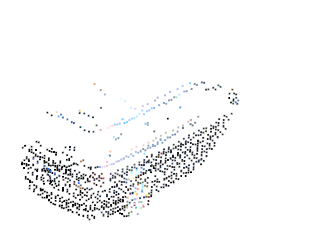}
	
	\includegraphics[width=0.091\textwidth]{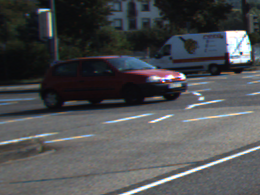}
	\includegraphics[width=0.091\textwidth]{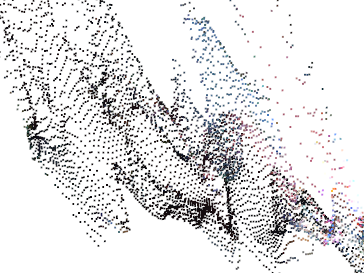}
	\includegraphics[width=0.091\textwidth]{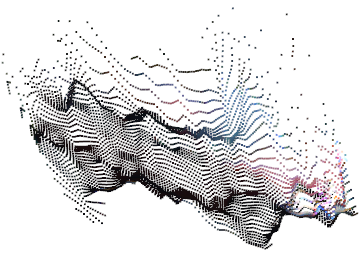}
	\includegraphics[width=0.091\textwidth]{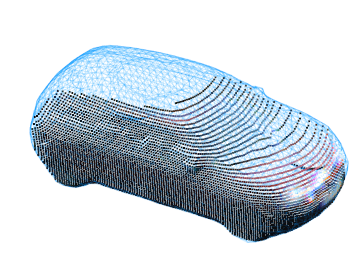}
	\includegraphics[width=0.091\textwidth]{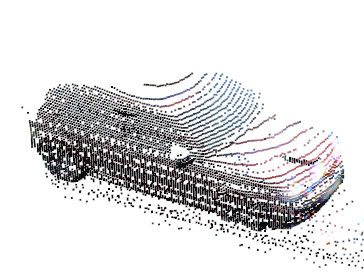}
	\begin{subfigure}[t]{.089\textwidth}
		\vspace{-2ex}
		\caption{Image}	
	\end{subfigure}
	\begin{subfigure}[t]{.096\textwidth}
		\vspace{-2ex}
		\caption{ELAS~\cite{geiger2010efficient}}	
	\end{subfigure}
	\begin{subfigure}[t]{.098\textwidth}
		\vspace{-2ex}
		\caption{PSMNet~\cite{chang2018pyramid}}	
	\end{subfigure}
	\begin{subfigure}[t]{.089\textwidth}
		\vspace{-2ex}
		\caption{Ours}	
	\end{subfigure}
	\begin{subfigure}[t]{.089\textwidth}
		\vspace{-2ex}
		\caption{GT}	
	\end{subfigure}
	
	\vspace{-1.5ex}
	\caption{Qualitative results on 3D shape refinement. We compare our method to a classical 
		(ELAS~\cite{geiger2010efficient}) and a SotA deep learning based 
		(PSMNet~\cite{chang2018pyramid}) stereo matching method.
	}\vspace{-3ex}
	\label{fig:eval_shape_quality}
\end{figure}

\begin{table*}[t]
	\begin{center}
		\begin{tabu}{|l|c c c|c c c|c c c|c c c|}
			\hline
			&
			\multicolumn{3}{c}{$\mathrm{AP_{bv}}$ (IoU=0.5)} & 
			\multicolumn{3}{c|}{$\mathrm{AP_{bv}}$ (IoU=0.7)} & 
			\multicolumn{3}{c}{$\mathrm{AP_{3D}}$ (IoU=0.5)} & 
			\multicolumn{3}{c|}{$\mathrm{AP_{3D}}$ (IoU=0.7)}\Tstrut\Bstrut \\
			\cline{2-13}
			Method & Easy & Mode & Hard & Easy & Mode & Hard & Easy & Mode & Hard & Easy & Mode 
			& 
			Hard\Tstrut\Bstrut \\
			\tabucline[1pt]{-}
			Mono3D \cite{cvpr16chen} & 11.70 & 9.62 & 9.32 & 2.06 & 1.91 & 1.39 & 9.55 & 7.72 & 
			7.23 & 0.62 & 0.75 & 0.76\Tstrut \\
			Mono3D + Ours & \textbf{23.53} & \textbf{16.54} & \textbf{15.30} & \textbf{5.21} & 
			\textbf{4.02} & \textbf{3.84} & \textbf{18.88} & \textbf{14.31} & \textbf{11.73} & 
			\textbf{2.61} & \textbf{2.09} & \textbf{2.17}\Bstrut \\
			\hline
			Deep3DBox \cite{mousavian20173d} & 29.99 & 23.74 & 18.81 & 9.96 & 7.69 & 5.29 & 26.94 & 
			20.51 & 15.85 & 5.82 & 4.08 & 3.83 \Tstrut\\
			Deep3DBox + Ours & \textbf{44.89} & \textbf{29.99} & \textbf{24.41} & \textbf{12.35} & 
			\textbf{8.88} & \textbf{7.49} & \textbf{38.40} & \textbf{25.39} & \textbf{20.02} 
			& \textbf{6.50} & \textbf{4.38} & \textbf{4.04}\Bstrut \\
			\hline
			3DOP \cite{3dopNIPS15} & 48.73 & 35.20 & 30.95 & 12.63 & 9.07 & 7.12 & 40.76 & 28.92 & 
			24.31 & 5.38 & 3.76 & 3.25\Tstrut \\
			3DOP + Ours & \textbf{59.40} & \textbf{39.43} & \textbf{33.54}& \textbf{19.98} & 
			\textbf{13.40} & \textbf{11.34} & \textbf{50.16} & \textbf{34.66} & \textbf{29.31} &  
			\textbf{11.38} & \textbf{7.36} & \textbf{6.34}\Bstrut \\
			\hline
			MLF \cite{xu2018multi} & 55.03 & 36.73 & 31.27 & 22.03 & 13.76 & 11.60 & 47.88 & 29.48 
			& 26.44 & 10.53 & 5.69 & 5.39\Tstrut \\
			MLF + Ours & \textbf{63.10} & \textbf{37.97} & \textbf{31.84}& \textbf{25.58} & 
			\textbf{15.25} & \textbf{11.97} & \textbf{55.12} & \textbf{34.78} & \textbf{29.44} &  
			\textbf{14.59} & \textbf{8.42} & \textbf{7.26}\Bstrut \\
			\hline
		\end{tabu}
	\end{center}
	\vspace{-2ex}
	\caption{Average precision of bird's eye view ($\mathrm{AP}_{bv}$) and 3D bounding boxes 
		($\mathrm{AP}_{3D}$), evaluated on the KITTI 3D Object validation set. Note that the KITTI 
		Object Benchmark updated its evaluation script in 2017 which causes some inconsistent 
		numbers in this table and in the original papers.}\vspace{-2ex}
	\label{tab:3d_pose}
\end{table*} 

\begin{figure*}[]
	\centering
	\includegraphics[width=0.34\textwidth,height=2cm]{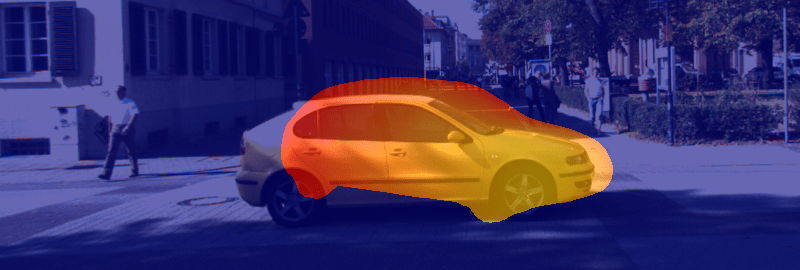}%
	\includegraphics[width=0.115\textwidth,height=2cm]{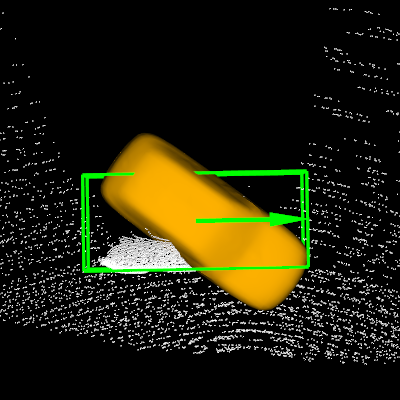}
	\includegraphics[width=0.34\textwidth,height=2cm]{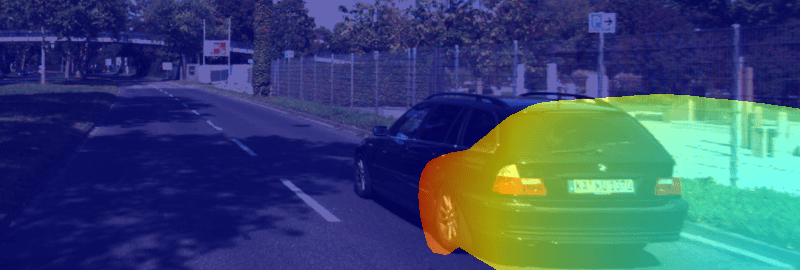}%
	\includegraphics[width=0.115\textwidth,height=2cm]{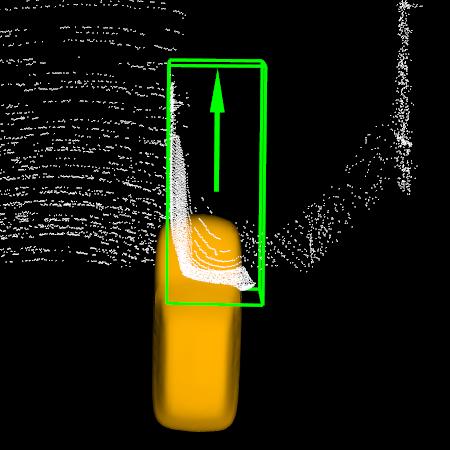}
	\includegraphics[width=0.34\textwidth,height=2cm]{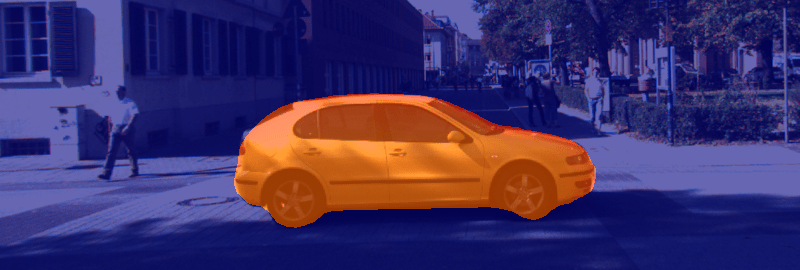}%
	\includegraphics[width=0.115\textwidth,height=2cm]{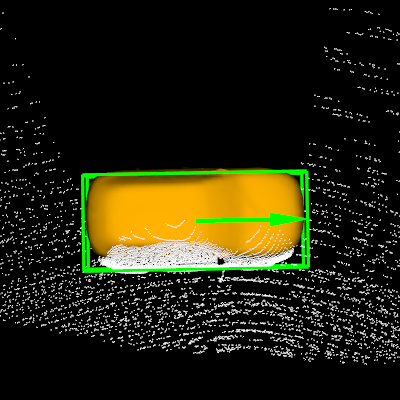}
	\includegraphics[width=0.34\textwidth,height=2cm]{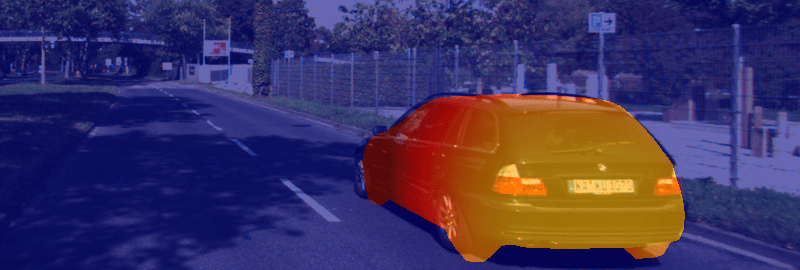}%
	\includegraphics[width=0.115\textwidth,height=2cm]{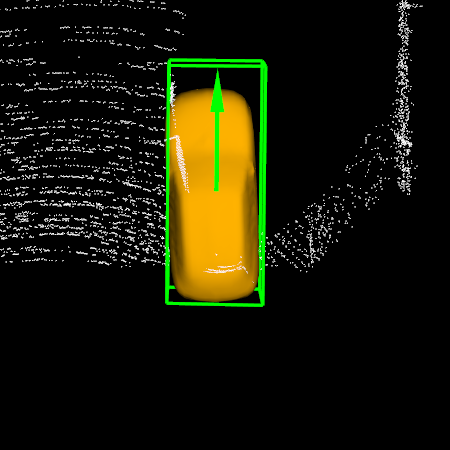}
	\caption{Qualitative results of 3D pose refinement. Each column shows the initial and the 
		optimized pose (overlapped with the input image and also in bird-eye view). GT poses are 
		denoted by green boxes. Note that point clouds are not used in our optimization.
	}\vspace{-0.5ex}
	\label{fig:eval_pose_quality}
\end{figure*}

\begin{figure*}[] 
	\centering
	
	\includegraphics[width=0.32\textwidth]{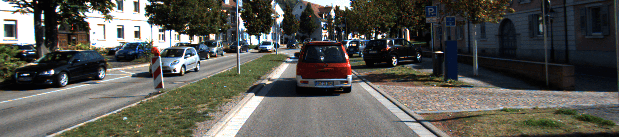}
	\includegraphics[width=0.32\textwidth]{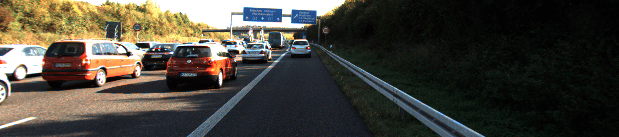}
	\includegraphics[width=0.32\textwidth]{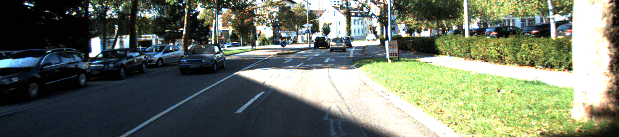}
	
	\includegraphics[width=0.32\textwidth]{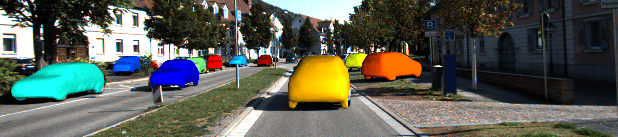}
	\includegraphics[width=0.32\textwidth]{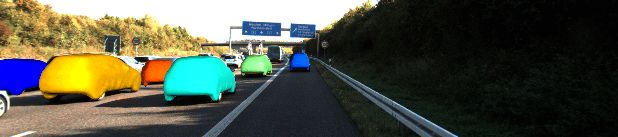}
	\includegraphics[width=0.32\textwidth]{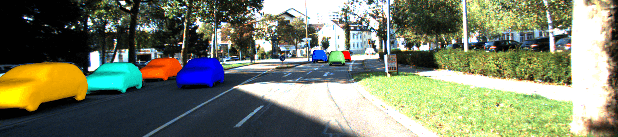}
	
	\vspace{1ex}
	
	\includegraphics[width=0.32\textwidth]{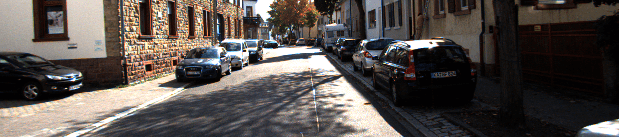}
	\includegraphics[width=0.32\textwidth]{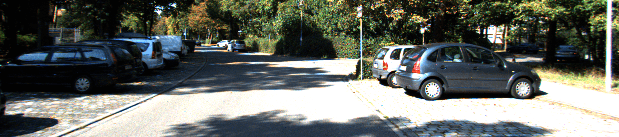}
	\includegraphics[width=0.32\textwidth]{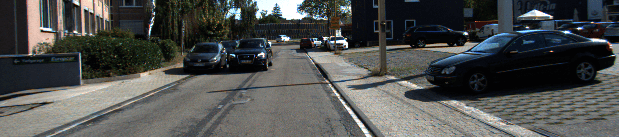}
	
	\includegraphics[width=0.32\textwidth]{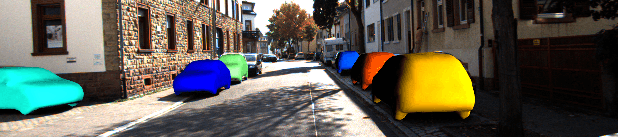}
	\includegraphics[width=0.32\textwidth]{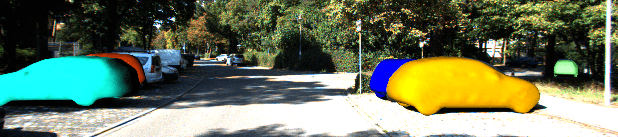}
	\includegraphics[width=0.32\textwidth]{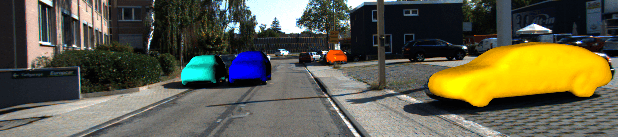}
	
	\caption{Qualitative results of 3D shape and pose estimation.
	}\vspace{-3ex}
	\label{fig:eval_quality}
\end{figure*}

We first compare our method to the previously proposed geometric 
approach~\cite{engelmann2016joint}, which fits the same PCA model to the point cloud estimated by 
the dense stereo reconstruction method ELAS~\cite{geiger2010efficient}. To our knowledge it is so far 
the only method that provides object shape evaluation. To be consistent 
with~\cite{ZhouICCV15,engelmann2017samp}, we measure $\mathit{completeness}$ (\% of 
ground-truth (GT) points with at least one estimated point within a certain distance $\tau$), 
$\mathit{accuracy}$ (\% of estimated points with at least one GT point within $\tau$) and $F_1$ 
score ($2\cdot \mathit{completeness} \cdot \mathit{accuracy} / 
(\mathit{completeness}+\mathit{accuracy})$) for the GT segments in the KITTI Stereo 
2015 benchmark. To more precisely measure the accuracy, we additionally compute the RMSE: For each 
GT point, we search within 
$\tau$ and compute the distance to the closest estimate point. The RMSE is only computed 
for those GT points with matched estimated points. We use 3DOP for pose initialization for 
\cite{engelmann2016joint} and our method. The four metrics wrt. different $\tau$ are displayed in 
Fig.~\ref{fig:eval_shape}, where our method outperforms the geometric approach in all the four 
evaluations\footnote{When comparing to Fig. 9 in~\cite{engelmann2017samp}, 
it is worth noting that the results there are obtained from a subset of KITTI Stereo 2015.}.
Compared to the laboratorial settings in~\cite{Sandhu2011_kernelsegposerecon, 
Prisacariu2012_segposerecon, DameCVPR13} where only one 
dominant car appears in the image, KITTI Stereo 2015 is much more challenging and contains 
cars with severe truncation, occlusion and at large distance. This explains the degraded performances 
of the geometric approach, as it lacks occlusion handling and also the stereo reconstruction gets 
drastically more noisy in faraway areas. 

In Fig.~\ref{fig:eval_shape_quality}, we qualitatively compare our method to 
ELAS~\cite{geiger2010efficient} and a recent deep learning based stereo reconstruction method 
PSMNet~\cite{chang2018pyramid}. Although deep learning with strong supervision has significantly 
improved the reconstruction quality, it still suffers at large distances. The results in 
Fig.~\ref{fig:eval_shape_quality} validate the idea of introducing shape priors into the pipeline. 
More qualitative results can be found in Fig.~\ref{fig:eval_quality}.

\subsection{Pose Refinement}
Same as above, we first compare our method to the geometric approach~\cite{engelmann2016joint}. 
Based on the GT provided by KITTI 3D Object, we compute the precision-recall curves for 3D bounding 
boxes for the three pre-defined difficulties. The results are shown in Fig~\ref{fig:eval_pose}, where our 
method delivers better 3D pose estimates for all the difficulties. In the categories Moderate and Hard, 
as many cars are occluded and only part of the object 3D points can be reconstructed, fitting the 
3D model to the incomplete point cloud would generally worsen the pose estimation. The qualitative 
results of our 3D pose refinement on two example images can be found in 
Fig.~\ref{fig:eval_pose_quality}.

In the next experiment we demonstrate the pose refinements of our method over 4 deep learning based 
3D detectors, namely Mono3D~\cite{cvpr16chen}, Deep3DBox~\cite{mousavian20173d}, 
3DOP~\cite{3dopNIPS15} and MLF~\cite{xu2018multi}. We use the validation splits provided 
by~\cite{3dopNIPS15,mousavian20173d} and compute the average precisions for bird's eye view 
($\mathrm{AP}_{bv}$) and 3D bounding boxes ($\mathrm{AP}_{3D}$). The results in 
Table~\ref{tab:3d_pose} shows that our method hugely boosts the performances of all the tested 
methods under all the settings, which demonstrates the effectiveness of our method on 3D 
pose refinement. Some qualitative results of our method in challenging real-world scenarios can be 
found 
in Fig.~\ref{fig:eval_quality}.

\section{Conclusions}
We propose a new approach for joint vehicle pose and shape estimation based on 
an energy function combining photometric and silhouette alignment. Our method delivers much more 
precise and useful information than the current 3D detectors that focus on estimating bounding 
boxes.
In our experiments we demonstrate superior performance over the previous geometric method in both 
pose and shape estimation. We also demonstrate that our approach can significantly boost the 
performance of learning-based 3D object detectors. In future work, we are planning to extend our 
approach to a local window of multiple frames and integrate it into a visual SLAM system.



\bibliographystyle{./IEEEtran}
\bibliography{IEEEabrv,bib}

\begin{thebibliography}{10}
\providecommand{\url}[1]{#1}
\csname url@rmstyle\endcsname
\providecommand{\newblock}{\relax}
\providecommand{\bibinfo}[2]{#2}
\providecommand\BIBentrySTDinterwordspacing{\spaceskip=0pt\relax}
\providecommand\BIBentryALTinterwordstretchfactor{4}
\providecommand\BIBentryALTinterwordspacing{\spaceskip=\fontdimen2\font plus
\BIBentryALTinterwordstretchfactor\fontdimen3\font minus
  \fontdimen4\font\relax}
\providecommand\BIBforeignlanguage[2]{{%
\expandafter\ifx\csname l@#1\endcsname\relax
\typeout{** WARNING: IEEEtran.bst: No hyphenation pattern has been}%
\typeout{** loaded for the language `#1'. Using the pattern for}%
\typeout{** the default language instead.}%
\else
\language=\csname l@#1\endcsname
\fi
#2}}

\bibitem{engelmann2016joint}
F.~Engelmann, J.~St{\"u}ckler, and B.~Leibe, ``Joint object pose estimation and
  shape reconstruction in urban street scenes using {3D} shape priors,'' in
  \emph{{German Conference on Pattern Recognition (GCPR)}}.\hskip 1em plus
  0.5em minus 0.4em\relax Springer, 2016, pp. 219--230.

\bibitem{engelmann2017samp}
------, ``{SAMP}: Shape and motion priors for {4D} vehicle reconstruction,'' in
  \emph{{Applications of Computer Vision (WACV), 2017 IEEE Winter Conference
  on}}.\hskip 1em plus 0.5em minus 0.4em\relax IEEE, 2017, pp. 400--408.

\bibitem{felzenszwalb2009object}
P.~F. Felzenszwalb, R.~B. Girshick, D.~McAllester, and D.~Ramanan, ``Object
  detection with discriminatively trained part-based models,'' \emph{IEEE
  transactions on pattern analysis and machine intelligence}, vol.~32, no.~9,
  pp. 1627--1645, 2009.

\bibitem{ren2015_fasterrcnn}
S.~Ren, K.~He, R.~Girshick, and J.~Sun, ``Faster {R-CNN}: Towards real-time
  object detection with region proposal networks,'' in \emph{Advances in Neural
  Information Processing Systems (NIPS)}, 2015, pp. 91--99.

\bibitem{dai2016r}
J.~Dai, Y.~Li, K.~He, and J.~Sun, ``{R-FCN}: Object detection via region-based
  fully convolutional networks,'' in \emph{Advances in neural information
  processing systems}, 2016, pp. 379--387.

\bibitem{dai2017deformable}
J.~Dai, H.~Qi, Y.~Xiong, Y.~Li, G.~Zhang, H.~Hu, and Y.~Wei, ``Deformable
  convolutional networks,'' in \emph{Proceedings of the IEEE international
  conference on computer vision}, 2017, pp. 764--773.

\bibitem{redmon2017_yolo9000}
J.~Redmon and A.~Farhadi, ``{YOLO9000}: Better, faster, stronger,'' in
  \emph{IEEE Conference on Computer Vision and Pattern Recognition (CVPR)},
  2017, pp. 6517--6525.

\bibitem{chen2017deeplab}
L.-C. Chen, G.~Papandreou, I.~Kokkinos, K.~Murphy, and A.~L. Yuille,
  ``{DeepLab}: Semantic image segmentation with deep convolutional nets, atrous
  convolution, and fully connected crfs,'' \emph{IEEE transactions on pattern
  analysis and machine intelligence}, vol.~40, no.~4, pp. 834--848, 2017.

\bibitem{he2017mask}
K.~He, G.~Gkioxari, P.~Doll{\'a}r, and R.~Girshick, ``{Mask R-CNN},'' in
  \emph{{Computer Vision (ICCV), 2017 IEEE International Conference on}}.\hskip
  1em plus 0.5em minus 0.4em\relax IEEE, 2017, pp. 2980--2988.

\bibitem{kirillov2019panoptic}
A.~Kirillov, K.~He, R.~Girshick, C.~Rother, and P.~Doll{\'a}r, ``Panoptic
  segmentation,'' in \emph{Proceedings of the IEEE Conference on Computer
  Vision and Pattern Recognition}, 2019, pp. 9404--9413.

\bibitem{satkin2013_3dnn}
S.~Satkin and M.~Hebert, ``{3DNN}: Viewpoint invariant {3D} geometry matching
  for scene understanding,'' in \emph{{IEEE} International Conference on
  Computer Vision, (ICCV)}, 2013, pp. 1873--1880.

\bibitem{3dopNIPS15}
X.~Chen, K.~Kundu, Y.~Zhu, A.~Berneshawi, H.~Ma, S.~Fidler, and R.~Urtasun,
  ``3d object proposals for accurate object class detection,'' in \emph{NIPS},
  2015.

\bibitem{shuran2014_3dslidingshapes}
S.~Song and J.~Xiao, ``Sliding shapes for 3d object detection in depth
  images,'' in \emph{European Conference on Computer Vision (ECCV)}, 2014, pp.
  634--651.

\bibitem{cvpr16chen}
X.~Chen, K.~Kundu, Z.~Zhang, H.~Ma, S.~Fidler, and R.~Urtasun, ``Monocular {3D}
  object detection for autonomous driving,'' in \emph{IEEE CVPR}, 2016.

\bibitem{mousavian20173d}
A.~Mousavian, D.~Anguelov, J.~Flynn, and J.~Kosecka, ``{3D} bounding box
  estimation using deep learning and geometry,'' in \emph{Proceedings of the
  IEEE Conference on Computer Vision and Pattern Recognition}, 2017, pp.
  7074--7082.

\bibitem{xu2018multi}
B.~Xu and Z.~Chen, ``Multi-level fusion based {3D} object detection from
  monocular images,'' in \emph{Proceedings of the IEEE Conference on Computer
  Vision and Pattern Recognition}, 2018, pp. 2345--2353.

\bibitem{licvpr2019}
P.~Li, X.~Chen, and S.~Shen, ``{Stereo R-CNN} based {3D} object detection for
  autonomous driving,'' in \emph{Proceedings of the IEEE Conference on Computer
  Vision and Pattern Recognition}, 2019.

\bibitem{wang2019pseudo}
Y.~Wang, W.-L. Chao, D.~Garg, B.~Hariharan, M.~Campbell, and K.~Q. Weinberger,
  ``{Pseudo-LiDAR} from visual depth estimation: Bridging the gap in {3D}
  object detection for autonomous driving,'' in \emph{Proceedings of the IEEE
  Conference on Computer Vision and Pattern Recognition}, 2019, pp. 8445--8453.

\bibitem{kundu20183d}
A.~Kundu, Y.~Li, and J.~M. Rehg, ``{3D-RCNN}: Instance-level {3D} object
  reconstruction via render-and-compare,'' in \emph{Proceedings of the IEEE
  Conference on Computer Vision and Pattern Recognition}, 2018, pp. 3559--3568.

\bibitem{salasmoreno2013_slampp}
R.~F. Salas-Moreno, R.~A. Newcombe, H.~Strasdat, P.~H.~J. Kelly, and A.~J.
  Davison, ``{SLAM++: Simultaneous Localisation and Mapping at the Level of
  Objects},'' in \emph{{2013 IEEE Conference on Computer Vision and Pattern
  Recognition}}, 2013, pp. 1352--1359.

\bibitem{Geiger2015GCPR}
A.~Geiger and C.~Wang, ``{Joint 3D Object and Layout Inference from a single
  RGB-D Image},'' in \emph{{German Conference on Pattern Recognition (GCPR)}},
  ser. Lecture Notes in Computer Science, vol. 9358.\hskip 1em plus 0.5em minus
  0.4em\relax Springer International Publishing, 2015, pp. 183--195.

\bibitem{Ortizcayon3DV16}
R.~Ortiz-Cayon, A.~Djelouah, F.~Massa, M.~Aubry, and G.~Drettakis, ``{Automatic
  3D Car Model Alignment for Mixed Image-Based Rendering},'' in
  \emph{International Conference on 3D Vision (3DV)}, 2016.

\bibitem{Sandhu2011_kernelsegposerecon}
R.~Sandhu, S.~Dambreville, A.~Yezzi, and A.~Tannenbaum, ``A nonrigid
  kernel-based framework for {2D-3D} pose estimation and {2D} image
  segmentation,'' \emph{IEEE Transactions on Pattern Analysis and Machine
  Intelligence}, vol.~33, no.~6, pp. 1098--1115, 2011.

\bibitem{Prisacariu2012_segposerecon}
V.~A. Prisacariu, A.~V. Segal, and I.~Reid, ``Simultaneous monocular {2D}
  segmentation, {3D} pose recovery and {3D} reconstruction,'' in \emph{Proc. of
  the Asian Conf. on Computer Vision (ACCV)}, 2013.

\bibitem{DameCVPR13}
A.~Dame, V.~A. Prisacariu, C.~Y. Ren, and I.~D. Reid, ``Dense reconstruction
  using {3D} object shape priors,'' in \emph{Proc. of the IEEE Int. Conf. on
  Computer Vision and Pattern Recognition (CVPR)}, 2013.

\bibitem{dambreville2008framework}
S.~Dambreville, Y.~Rathi, and A.~Tannenbaum, ``A framework for image
  segmentation using shape models and kernel space shape priors,'' \emph{{IEEE
  Transactions on Pattern Analysis and Machine Intelligence (TPAMI)}}, vol.~30,
  no.~8, pp. 1385--1399, 2008.

\bibitem{sandhu2011nonrigid}
R.~Sandhu, S.~Dambreville, A.~Yezzi, and A.~Tannenbaum, ``A nonrigid
  kernel-based framework for {2D-3D} pose estimation and {2D} image
  segmentation,'' \emph{{IEEE Transactions on Pattern Analysis and Machine
  Intelligence (TPAMI)}}, vol.~33, no.~6, pp. 1098--1115, 2011.

\bibitem{prisacariu2011nonlinear}
V.~A. Prisacariu and I.~Reid, ``Nonlinear shape manifolds as shape priors in
  level set segmentation and tracking,'' in \emph{{Computer Vision and Pattern
  Recognition (CVPR), 2011 IEEE Conference on}}.\hskip 1em plus 0.5em minus
  0.4em\relax IEEE, 2011, pp. 2185--2192.

\bibitem{prisacariu2011shared}
------, ``Shared shape spaces,'' in \emph{{International Conference on Computer
  Vision (ICCV)}}.\hskip 1em plus 0.5em minus 0.4em\relax IEEE, 2011, pp.
  2587--2594.

\bibitem{prisacariu2012simultaneous}
V.~A. Prisacariu, A.~V. Segal, and I.~Reid, ``Simultaneous monocular {2D}
  segmentation, {3D} pose recovery and {3D} reconstruction,'' in \emph{{Asian
  Conference on Computer Vision}}.\hskip 1em plus 0.5em minus 0.4em\relax
  Springer, 2012, pp. 593--606.

\bibitem{zheng2015object}
S.~Zheng, V.~A. Prisacariu, M.~Averkiou, M.-M. Cheng, N.~J. Mitra, J.~Shotton,
  P.~H. Torr, and C.~Rother, ``Object proposals estimation in depth image using
  compact 3d shape manifolds,'' in \emph{{German Conference on Pattern
  Recognition (GCPR)}}.\hskip 1em plus 0.5em minus 0.4em\relax Springer, 2015,
  pp. 196--208.

\bibitem{engel2013semi}
J.~Engel, J.~Sturm, and D.~Cremers, ``Semi-dense visual odometry for a
  monocular camera,'' in \emph{{Computer Vision (ICCV), 2013 IEEE International
  Conference on}}.\hskip 1em plus 0.5em minus 0.4em\relax IEEE, 2013, pp.
  1449--1456.

\bibitem{engel2014lsd}
J.~Engel, T.~Sch{\"o}ps, and D.~Cremers, ``{LSD-SLAM}: Large-scale direct
  monocular {SLAM},'' in \emph{{European Conference on Computer Vision
  (ECCV)}}.\hskip 1em plus 0.5em minus 0.4em\relax Springer, 2014, pp.
  834--849.

\bibitem{wang2017stereo}
R.~Wang, M.~Schw{\"o}rer, and D.~Cremers, ``{Stereo DSO}: Large-scale direct
  sparse visual odometry with stereo cameras,'' in \emph{{International
  Conference on Computer Vision (ICCV), Venice, Italy}}, 2017.

\bibitem{engel2018direct}
J.~Engel, V.~Koltun, and D.~Cremers, ``Direct sparse odometry,'' \emph{IEEE
  Transactions on Pattern Analysis and Machine Intelligence (TPAMI)}, vol.~40,
  no.~3, pp. 611--625, 2018.

\bibitem{yang2018deep}
N.~Yang, R.~Wang, J.~Stuckler, and D.~Cremers, ``Deep virtual stereo odometry:
  Leveraging deep depth prediction for monocular direct sparse odometry,'' in
  \emph{Proceedings of the European Conference on Computer Vision (ECCV)},
  2018, pp. 817--833.

\bibitem{gao2018ldso}
X.~Gao, R.~Wang, N.~Demmel, and D.~Cremers, ``{LDSO}: Direct sparse odometry
  with loop closure,'' in \emph{2018 IEEE/RSJ International Conference on
  Intelligent Robots and Systems (IROS)}.\hskip 1em plus 0.5em minus
  0.4em\relax IEEE, 2018, pp. 2198--2204.

\bibitem{zhu2018object}
R.~Zhu, C.~Wang, C.-H. Lin, Z.~Wang, and S.~Lucey, ``Object-centric photometric
  bundle adjustment with deep shape prior,'' in \emph{2018 IEEE Winter
  Conference on Applications of Computer Vision (WACV)}.\hskip 1em plus 0.5em
  minus 0.4em\relax IEEE, 2018, pp. 894--902.

\bibitem{Menze2015CVPR}
M.~Menze and A.~Geiger, ``Object scene flow for autonomous vehicles,'' in
  \emph{{Conference on Computer Vision and Pattern Recognition (CVPR)}}, 2015.

\bibitem{Geiger2012CVPR}
A.~Geiger, P.~Lenz, and R.~Urtasun, ``Are we ready for autonomous driving? the
  {KITTI} vision benchmark suite,'' in \emph{{Conference on Computer Vision and
  Pattern Recognition (CVPR)}}, 2012.

\bibitem{geiger2010efficient}
A.~Geiger, M.~Roser, and R.~Urtasun, ``Efficient large-scale stereo matching,''
  in \emph{{Asian conference on computer vision}}.\hskip 1em plus 0.5em minus
  0.4em\relax Springer, 2010, pp. 25--38.

\bibitem{chang2018pyramid}
J.-R. Chang and Y.-S. Chen, ``Pyramid stereo matching network,'' in
  \emph{Proceedings of the IEEE Conference on Computer Vision and Pattern
  Recognition}, 2018, pp. 5410--5418.

\bibitem{ZhouICCV15}
C.~Zhou, F.~G{\"u}ney, Y.~Wang, and A.~Geiger, ``Exploiting object similarity
  in {3D} reconstruction,'' in \emph{Proc. of the IEEE Int. Conf. on Computer
  Vision (ICCV)}, 2015.

\end{thebibliography}

\end{document}


\maketitle
\thispagestyle{empty}
\pagestyle{empty}

\begin{abstract}	
In this \textbf{supplementary document}, we first show the shape variations of the 
adopted PCA model in Sec.~\ref{shape_pca}. 
In Sec.~\ref{full_jacobian}, the full derivations of the analytical Jacobians of all the residuals 
defined in the main paper are presented.
Lastly in Sec.~\ref{more_res}, we show more results on the KITTI dataset, which qualitatively 
demonstrate the ability of our method to recover the 3D poses and shapes of cars in challenging 
real-world environments. 
Apart from this document, a video showing how our method works on some selected stereo frames can 
be found on the project page \url{https://vision.in.tum.de/research/vslam/direct-shape}.

\end{abstract}

\section{Shape Variations of PCA Model}
\label{shape_pca}
\begin{figure*}
	\centering
	\begin{subfigure}[c]{.90\textwidth}
		\includegraphics[width=0.25\textwidth]{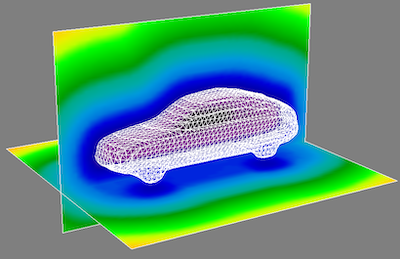}%
		\includegraphics[width=0.25\textwidth]{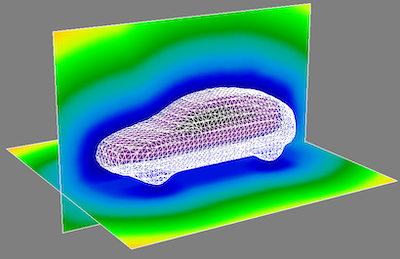}%
		\includegraphics[width=0.25\textwidth]{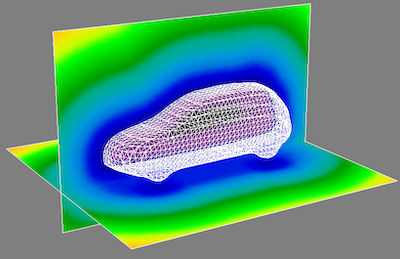}%
		\includegraphics[width=0.25\textwidth]{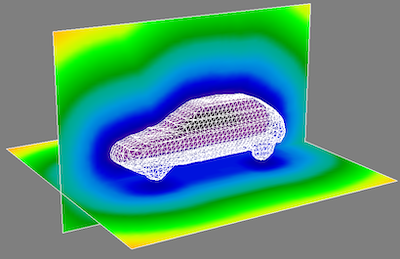}
		\includegraphics[width=0.25\textwidth]{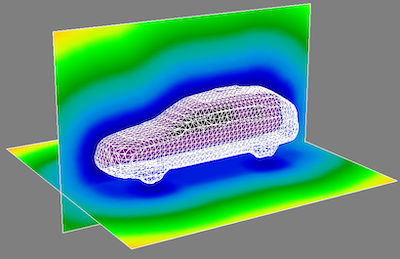}%
		\includegraphics[width=0.25\textwidth]{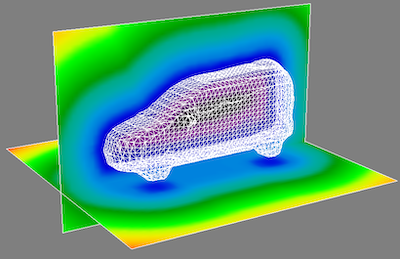}%
		\includegraphics[width=0.25\textwidth]{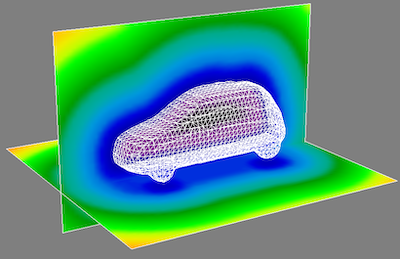}%
		\includegraphics[width=0.25\textwidth]{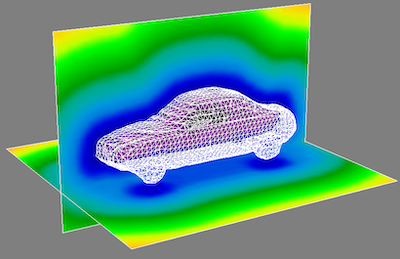}
		\includegraphics[width=0.25\textwidth]{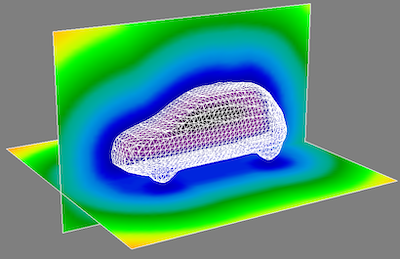}%
		\includegraphics[width=0.25\textwidth]{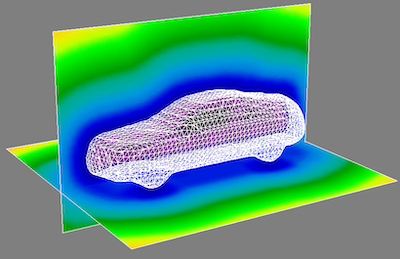}%
		\includegraphics[width=0.25\textwidth]{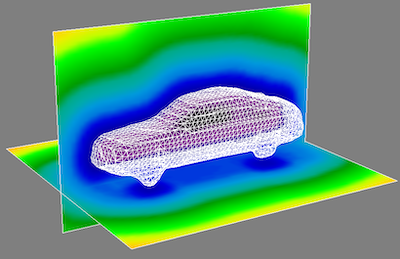}%
		\includegraphics[width=0.25\textwidth]{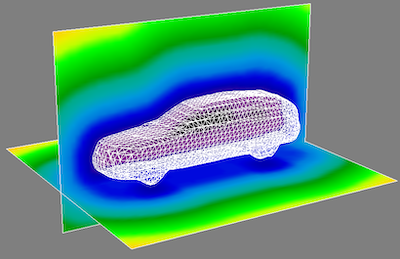}
	\end{subfigure}
	\begin{subfigure}[c]{.09\textwidth}
		\includegraphics[width=0.8\textwidth]{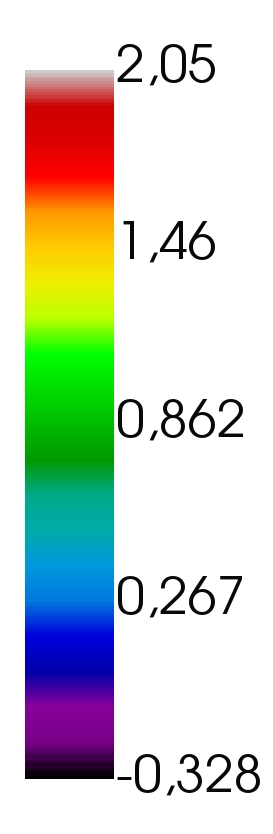}
	\end{subfigure}
	\caption{Shape variations of PCA model with color coded SDF.}
	\label{sdfs}
\end{figure*}

\begin{figure*}[!]
	\centering
	\includegraphics[width=0.96\textwidth]{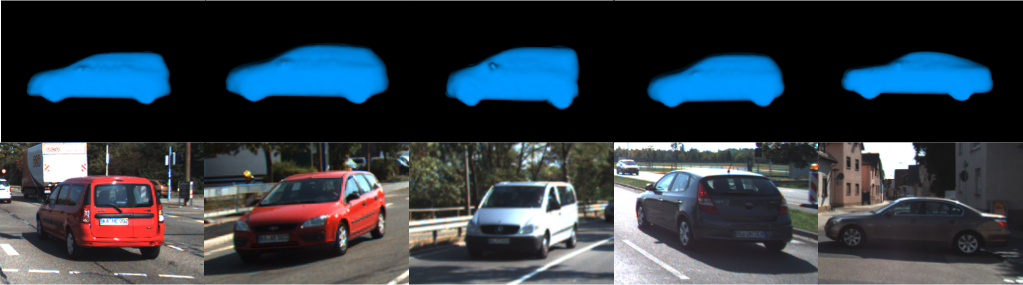}
	\caption{Qualitative results on the shape estimation.}\vspace{-2ex}
	\label{fig:shape_variation}
\end{figure*}

To demonstrate that our PCA model can deform and fit properly to a variety of car shapes, we first fit it 
to 12 selected vehicles from the CAD samples which we used to extract the PCA model. The shapes 
together with the color coded signed distance function (SDF) are shown in Fig.~\ref{sdfs}. An 
animation showing the different car shapes by modifying the shape coefficients $\mathbf{z}$ can be 
found on the project page.
We further show some real-world examples in Fig.~\ref{fig:shape_variation}, where the PCA model is 
optimized using our approach to fit the corresponding cars in the second row. We claim that 
although the adopted PCA shape embedding is 
a simple linear model, it works nicely for object categories like cars. 

\section{Full Derivations of Jacobians}
\label{full_jacobian}
\subsection{Jacobian of Silhouette Alignment Residual} 
As the relative transformation between the left and right cameras are considered to be fixed in 
this work, the Jacobians of $r_{silh}^{l}$ and $r_{silh}^{r}$ are the same and we will omit the 
superscript in the following. As shown in Eq. 2 in the main paper, the silhouette alignment 
residual of pixel $\mathbf{p}$ is defined as
\begin{equation}
r_{silh}(\mathbf{p}) =  
-\text{log} \big( \underbrace{\pi(\mathbf{\Phi}, \mathbf{p}) p_{fg}(\mathbf{p}) 
	+ (1 - \pi(\mathbf{\Phi}, \mathbf{p})) p_{bg}(\mathbf{p})}_{\coloneqq  
	A(\pi)} 
\big),
\end{equation}
thus using chain rule its Jacobian with respect to the pose and shape parameters 
$[\boldsymbol{\xi}_{c}^{o}; \mathbf{z}]$ can be factorized to
\begin{align}
\mathbf{J}_{silh} & = 
\frac{\partial r_{silh}(\mathbf{p})}{\partial 
	[\boldsymbol{\xi}_{c}^{o}; \mathbf{z}]} \\ & = 
-\frac{\partial log(A(\pi))}{\partial A(\pi)} 
\frac{\partial A(\pi)}{\partial \pi} \frac{\partial \pi(\mathbf{\Phi}, 
	\mathbf{p})}{\partial \mathbf{\Phi}} \frac{\partial 
	\mathbf{\Phi}}{\partial [\boldsymbol{\xi}_{c}^{o}; \mathbf{z}]},
\end{align}
where
\begin{align}
\frac{\partial log(A(\pi))}{\partial A(\pi)}  = 
\frac{1}{A(\pi)},
\label{eq:dlog_dA}
\end{align}
\begin{align}
\frac{\partial A(\pi)}{\partial \pi}  = 
p_{fg}(\mathbf{p}) - p_{bg}(\mathbf{p}).
\label{eq:dA_dpi}
\end{align}
Recall that the shape embedding projection function $\pi(\mathbf{\Phi}, \mathbf{p})$ is defined as
\begin{equation}
\pi(\mathbf{\Phi}, \mathbf{p}) = 1 - \prod_{\mathbf{X}_{o}} 
\frac{1}{e^{\mathbf{\Phi}(\mathbf{X}_{o})\zeta} + 1},
\end{equation}
to make it easier to derive its Jacobian, we convert the multiplications in $\pi(\mathbf{\Phi}, 
\mathbf{p})$ to summations by reformulating it to
\begin{equation}
\pi(\mathbf{\Phi}, \mathbf{p}) = 
1 - \text{exp}\bigg(\underbrace{\sum_{\mathbf{X}_{o}}  
	\text{log}\bigg(\overbrace{\frac{1} 
		{e^{\mathbf{\Phi}(\mathbf{X}_{o})\zeta}
			+ 1}}^{\coloneqq C(\mathbf{\Phi})}\bigg)}_{\coloneqq B(\mathbf{\Phi})} \bigg).
\end{equation}
Therefore, 
\begin{align}
\frac{\partial \pi(\mathbf{\Phi}, \mathbf{p})}{\partial \mathbf{\Phi}} = 
-\text{exp}(B(\mathbf{\Phi})) \sum_{\mathbf{X}_{o}} \frac{1}{C(\mathbf{\Phi})} 
\frac{\partial C(\mathbf{\Phi})}{\partial \mathbf{\Phi}},
\label{eq:dpi_dPhi}
\end{align}
where
\begin{align}
\frac{\partial C(\mathbf{\Phi})}{\partial \mathbf{\Phi}} & = 
\frac{\partial (\frac{1}{e^{\mathbf{\Phi}(\mathbf{X}_{o})\zeta} + 
		1})}{\partial \mathbf{\Phi}} \\ & = 
(-1) \frac{e^{\mathbf{\Phi}(\mathbf{X}_{o})\zeta} 
	}{(e^{\mathbf{\Phi}(\mathbf{X}_{o})\zeta} + 
	1)^2}\zeta \\ & = 
-\frac{\zeta e^{\mathbf{\Phi}(\mathbf{X}_{o})\zeta} 
}{(e^{\mathbf{\Phi}(\mathbf{X}_{o})\zeta} + 
	1)^2}.
\label{eq:dC_dPhi}
\end{align}
The remaining part to derive is $\partial \mathbf{\Phi}(\mathbf{X}_{o}) / \partial 
[\boldsymbol{\xi}_{c}^{o}; \mathbf{z}]$. 
As $\mathbf{\Phi}(\mathbf{X}_{o}) = \mathbf{V(\mathbf{X}_{o})z} + \mathbf{\Phi}_{mean} = 
\sum_{k=1}^{K}\mathbf{v}_{k}(\mathbf{X}_{o})z_k + \mathbf{\Phi}_{mean}$, we have
\begin{equation}
\label{eq:dPhi_dz}
\frac{\partial \mathbf{\Phi}(\mathbf{X}_{o})}{\partial \mathbf{z}} = 
[\mathbf{v}_{1}(\mathbf{X}_{o}), 
\mathbf{v}_{2}(\mathbf{X}_{o}),...,\mathbf{v}_{K}(\mathbf{X}_{o})].
\end{equation}
To derive $\partial 
\mathbf{\Phi}(\mathbf{X}_{o}) / \partial \boldsymbol{\xi}_{c}^{o}$, we first 
compute the coordinate for $\mathbf{X}_{o}$ in the camera coordinate system as 
$\mathbf{X}_{c}$, so we have $\mathbf{X}_{o} = 
\text{exp}\left(\hat{\boldsymbol{\xi}_{c}^{o}}\right)\mathbf{X}_{c}$, where 
$\text{exp}(\hat{\cdot})$ 
is the exponential map that maps the twist coordinate to $\text{SE(3)}$. The 
remaining part of the Jacobian is then computed as
\begin{equation}
\label{eq:dPhi_dxi}
\frac{\partial \mathbf{\Phi}(\mathbf{X}_{o})}{\partial 
	\boldsymbol{\xi}_{c}^{o}} = \nabla \mathbf{\Phi} \bigg 
\rvert_{\mathbf{X}_{o}} \frac{\partial \mathbf{X}_{o}}{\partial 
	\boldsymbol{\xi}_{c}^{o}},
\end{equation}
\begin{equation}
\label{eq:dx_dxi}
\frac{\partial \mathbf{X}_{o}}{\partial 
	\boldsymbol{\xi}_{c}^{o}} =
\frac{\partial \text{exp}\left(\hat{\boldsymbol{\xi}_{c}^{o}}\right)}{\partial 
	\boldsymbol{\xi}_{c}^{o}}\bigg \rvert_{\boldsymbol{\xi}_{c}^{o}} 
\mathbf{X}_{c} = 
\frac{\partial \text{exp}(\hat{\delta 
		\boldsymbol{\xi}})}{\partial (\delta \boldsymbol{\xi})} \bigg 
\rvert_{\mathbf{0}} 
\text{exp}(\hat{\boldsymbol{\xi}_{c}^{o}}) \mathbf{X}_{c},
\end{equation}
where $\nabla \mathbf{\Phi}$ is the spatial gradient of $\mathbf{\Phi}$,
$\delta \boldsymbol{\xi}$ is a small increment in $\mathfrak{se}(3)$ and 
is applied with the exponential map to the left hand side of the pose estimate. 
The closed form solution for $\partial \text{exp}(\hat{\delta \boldsymbol{\xi}}) / \partial (\delta 
\boldsymbol{\xi})$ near $\delta \boldsymbol{\xi} = \mathbf{0}$ can be obtained using the 
infinitesimal generators of $\text{SE(3)}$ (please refer Eq.~\ref{eq:generator1} and 
\ref{eq:generator2}).

Depending on the derivations of the specific derivatives above, the full Jacobian of the silhouette 
alignment residual can be computed by combining Eq.~\ref{eq:dlog_dA}, \ref{eq:dA_dpi}, 
\ref{eq:dpi_dPhi}, \ref{eq:dC_dPhi} and Eq.~\ref{eq:dPhi_dz}, \ref{eq:dPhi_dxi}, \ref{eq:dx_dxi}.

\subsection{Jacobian of the Photometric Consistency Residual} 
As defined in the main paper, the photometric consistency residual of pixel $\mathbf{p}$ is
\begin{equation}
r_{photo}(\mathbf{p}) =
\mathbf{I}_{r}\big(\underbrace{\Pi_{c}(\mathbf{R}_{l}^{r} 
	\Pi_{c}^{-1}(\mathbf{p}, 
	d_{\mathbf{p}}) + \mathbf{t}_{l}^{r})}_{\coloneqq warp(\mathbf{p}, 
	d_{\mathbf{p}})} \big) - \mathbf{I}_{l}\big(\mathbf{p}\big),
\end{equation}
where the pose and shape parameters $[\boldsymbol{\xi}_{c}^{o}; \mathbf{z}]$ only appear in 
$d_{\mathbf{p}}$. Using chain rule the Jacobian with respect to the pose and shape parameters can 
be factorized to 
\begin{align}
\mathbf{J}_{photo} & = 	\frac{\partial r_{photo}(\mathbf{p})}{\partial 
	[\boldsymbol{\xi}_{c}^{o}; \mathbf{z}]} \\ & = 
\nabla \mathbf{I}_r(warp(\mathbf{p}, d_{\mathbf{p}})) \frac{\partial warp(\mathbf{p}, 
	d_{\mathbf{p}})}{\partial d_{\mathbf{p}}} \frac{\partial 
	d_{\mathbf{p}}}{\partial [\boldsymbol{\xi}_{c}^{o}; \mathbf{z}]},
\end{align}
where $warp(\mathbf{p}, d_{\mathbf{p}}) = \Pi_{c}(\mathbf{R}_{l}^{r} 
\Pi_{c}^{-1}(\mathbf{p}, d_{\mathbf{p}}) + \mathbf{t}_{l}^{r})$ is the pixel 
warping function from the left image to the right image, $\nabla 
\mathbf{I}_r(warp(\mathbf{p}, d_{\mathbf{p}}))$ is the image gradient of the right image 
at the warped pixel location $warp(\mathbf{p}, d_{\mathbf{p}})$. In the following we derive 
$\partial warp(\mathbf{p}, d_{\mathbf{p}}) / \partial d_{\mathbf{p}}$ and $\partial d_{\mathbf{p}} 
/ \partial [\boldsymbol{\xi}_{c}^{o}; \mathbf{z}]$ successively.

Denoting the 3D coordinates of $\mathbf{p}$ in the left and the right camera 
coordinate systems as $\mathbf{X}_l$ and $\mathbf{X}_r$, we have
\begin{align}
warp(\mathbf{p}, d_{\mathbf{p}}) & = 
\Pi_{c}(\underbrace{\mathbf{R}_{l}^{r} 
	\overbrace{\Pi_{c}^{-1}(\mathbf{p}, d_{\mathbf{p}})}^{\mathbf{X}_l} + 
	\mathbf{t}_{l}^{r}}_{\mathbf{X}_r}),
\end{align}
\begin{align}
\mathbf{X}_l & =
d_{\mathbf{p}}\mathbf{K}^{-1}[\mathbf{p}(u), \mathbf{p}(v), 1]^{\top},
\end{align}
\begin{align}
\mathbf{X}_r & =
\mathbf{R}_{l}^{r}\mathbf{X}_l + \mathbf{t}_{l}^{r},\\ & =
d_{\mathbf{p}}\underbrace{\mathbf{R}_{l}^{r}\mathbf{K}^{-1}[\mathbf{p}(u), 
	\mathbf{p}(v), 1]^{\top}}_{\coloneqq \mathbf{v} = [\mathbf{v}(x), 
	\mathbf{v}(y), \mathbf{v}(z)]^{\top}} + \mathbf{t}_{l}^{r}, \\ & =
d_{\mathbf{p}}\mathbf{v} + \mathbf{t}_{l}^{r}, 
\end{align}
\begin{align}
\Pi_{c}(\mathbf{X}_r) & = 
\begin{bmatrix}
f_u & 0 & c_u \\
0 & f_v & c_v
\end{bmatrix}
\begin{bmatrix}
\frac{\mathbf{X}_r(x)}{\mathbf{X}_r(z)} \\
\frac{\mathbf{X}_r(y)}{\mathbf{X}_r(z)} \\
1
\end{bmatrix} \\ & =
\begin{bmatrix}
f_u\frac{\mathbf{X}_r(x)}{\mathbf{X}_r(z)} + c_u \\
f_v\frac{\mathbf{X}_r(y)}{\mathbf{X}_r(z)} + c_v, 
\end{bmatrix},
\end{align}
where $\mathbf{K} = [f_u, 0, c_u; 0, f_v, c_v; 0, 0, 1]$ is the camera 
intrinsic matrix. $\partial warp(\mathbf{p}, d_{\mathbf{p}}) / \partial 
d_{\mathbf{p}}$ therefore can be computed as
\begin{align}
\frac{\partial warp(\mathbf{p}, d_{\mathbf{p}})}{\partial d_{\mathbf{p}}} & = 
\begin{bmatrix}
f_u \frac{\partial \frac{\mathbf{X}_r(x)}{\mathbf{X}_r(z)}}{d_{\mathbf{p}}} \\
f_v \frac{\partial \frac{\mathbf{X}_r(y)}{\mathbf{X}_r(z)}}{d_{\mathbf{p}}} 
\end{bmatrix} \\ & =
\begin{bmatrix}
f_u \frac{\frac{\partial \mathbf{X}_r(x)}{\partial d_{\mathbf{p}}} 
	\mathbf{X}_r(z) - \mathbf{X}_r(x)\frac{\partial \mathbf{X}_r(z)}{\partial 
		d_{\mathbf{p}}}}{\mathbf{X}_r^2(z)} \\
f_v \frac{\frac{\partial \mathbf{X}_r(y)}{\partial d_{\mathbf{p}}} 
	\mathbf{X}_r(z) - \mathbf{X}_r(y)\frac{\partial \mathbf{X}_r(z)}{\partial 
		d_{\mathbf{p}}}}{\mathbf{X}_r^2(z)}
\end{bmatrix} \\ & = 
\begin{bmatrix}
f_u \frac{\mathbf{v}(x)\mathbf{X}_r(z) - 
	\mathbf{X}_r(x)\mathbf{v}(z)}{\mathbf{X}_r^2(z)} \\
f_v \frac{\mathbf{v}(y)\mathbf{X}_r(z) - 
	\mathbf{X}_r(y)\mathbf{v}(z)}{\mathbf{X}_r^2(z)} \\
\end{bmatrix}.
\end{align}

\begin{figure}[!]
	\centering
	\includegraphics[width=0.25\textwidth]{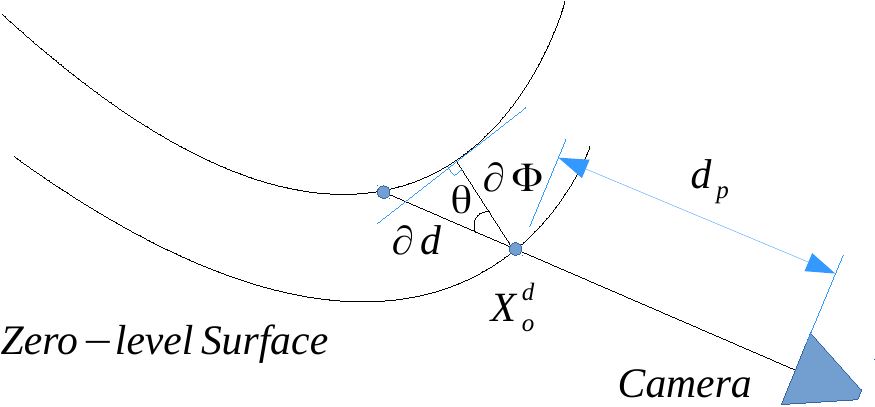}
	\vspace{-0.5ex}
	\caption{Deriving the Jacobian of the depth wrt. the SDF value.}\vspace{-3ex}
	\label{fig:tsdf_derivative}
\end{figure}

To compute $\partial d_{\mathbf{p}} / \partial [\boldsymbol{\xi}_{c}^{o}; \mathbf{z}]$, we first 
compute the 3D coordinate of the intersecting point of the ray and the zero-level surface based on 
$d_{\mathbf{p}}$ obtained by ray-casting, then transform it from the camera coordinate system to 
the object coordinate system and denote it as $\mathbf{X}_o^d$. The Jacobian with respect to the 
shape encoding vector is then computed as
\begin{equation}
\frac{\partial d_{\mathbf{p}}}{\partial \mathbf{z}} = \frac{\partial 
	d_{\mathbf{p}}}{\partial \mathbf{\Phi}} \bigg 
\rvert_{\mathbf{\Phi}(\mathbf{X}_o^d)} \frac{\partial \mathbf{\Phi}}{\partial 
	\mathbf{z}} \bigg \rvert_{\mathbf{z}},	
\end{equation}
where $\partial \mathbf{\Phi} / \partial \mathbf{z}$ can be computed similarly as in 
Eq.~\ref{eq:dPhi_dz}, the derivation of $\partial d_{\mathbf{p}} / \partial \mathbf{\Phi}$ is 
illustrated in Fig.~\ref{fig:tsdf_derivative}. At the intersecting point $\mathbf{X}^{d}_{o}$, the 
change of the depth along the ray $\partial d$ is approximately proportional to the change of the 
SDF value $\delta \mathbf{\Phi}$ by a factor of $1/\text{cos}(\theta)$ where $\theta$ is the 
angle between the ray and the surface normal. Taking the sign into account we have
\begin{equation}
\label{eq:dd_dPhi}
\frac{\partial d_{\mathbf{p}}}{\partial \mathbf{\Phi}}\bigg 
\rvert_{\mathbf{\Phi}(\mathbf{X}_o^d)} = 
-\frac{1}{\text{cos}(\theta)}.
\end{equation}
The Jacobian with respect to $\boldsymbol \xi^{o}_{c}$ can be factorized to 
\begin{equation}
\frac{\partial d_{\mathbf{p}}}{\partial \boldsymbol{\xi}^{o}_{c}} = 
\frac{\partial d_{\mathbf{p}}}{\partial \mathbf{\Phi}} \bigg 
\rvert_{\mathbf{\Phi}(\mathbf{X}_o^d)} \nabla \mathbf{\Phi} \bigg 
\rvert_{\mathbf{X}_{o}^{d}} \frac{\partial \mathbf{X}_{o}^{d}}{\partial 
	\boldsymbol{\xi}_{c}^{o}} \bigg \rvert_{\boldsymbol{\xi}_{c}^{o}},
\end{equation}
which can be computed according to Eq.~\ref{eq:dd_dPhi} and~\ref{eq:dx_dxi}.

\subsection{Jacobian of Prior Residuals} 
Based on the energy terms defined in the Eq. 7-9 in the main paper, we define the residuals of the 
priors 
on the shape and pose parameters as
\begin{align}
r_{shape}^{i} & =
\frac{z_i}{\sigma_i}, \quad i = 1, 2,..., K \label{eq:shape}\\
r_{trans} & = 
\mathbf{t}_o^c(y) - g(\mathbf{t}_o^c(x, z))(y), \\
r_{rot} & =
1 -  (\mathbf{R}_{o}^{c}[0, -1, 0]^{\top})^{\top} \mathbf{n}_{g}.
\end{align}

\subsubsection{Jacobian of Shape Prior Residuals.} Based on Eq.~\ref{eq:shape} 
we have
\begin{align}
\frac{\partial r_{shape}^{i}}{\partial \boldsymbol{\xi}_{c}^{o}} & = 
\mathbf{0},\\
\frac{\partial r_{shape}^{i}}{\partial \mathbf{z}} & = [0,...,0, 
\frac{1}{\sigma_i}, 0, 
..., 0].
\end{align}
\subsubsection{Jacobian of Translation Prior Residuals.} Denoting the equation 
for the ground plane as $\mathbf{n}_g(x)x + \mathbf{n}_g(y)y + \mathbf{n}_g(z)z 
+ d = 0$ with $\mathbf{n}_g$ the plane normal vector and $d$ a constant, the 
height of the ground plane at $\mathbf{t}_o^c(x, z)$ is
\begin{align}
g(\mathbf{t}_o^c(x, z))(y) = -\frac{\mathbf{n}_g(x)\mathbf{t}_o^c(x) + 
	\mathbf{n}_g(z)\mathbf{t}_o^c(z) + d}{\mathbf{n}_g(y)},
\end{align}
thus
\begin{align}
r_{trans} = \mathbf{t}_o^c(y) + \frac{\mathbf{n}_g(x)\mathbf{t}_o^c(x) + 
	\mathbf{n}_g(z)\mathbf{t}_o^c(z) + d}{\mathbf{n}_g(y)}.
\end{align}
Its Jacobian with respect to $\boldsymbol{\xi}_{c}^{o}$ then can be computed as
\begin{align}
\frac{\partial r_{trans}}{\partial \boldsymbol{\xi}_{c}^{o}} & =
\frac{\partial r_{trans}}{\partial \mathbf{t}_o^c}\frac{\partial 
	\mathbf{t}_o^c}{\partial \boldsymbol{\xi}_{c}^{o}} \\ & =
[\frac{\mathbf{n}_g(x)}{\mathbf{n}_g(y)}, 1, 
\frac{\mathbf{n}_g(z)}{\mathbf{n}_g(y)}]\frac{\partial 
	\mathbf{t}_o^c}{\partial \boldsymbol{\xi}_{c}^{o}},
\end{align}
where the last term can be computed as
\begin{align}
\frac{\partial \mathbf{t}_o^c}{\partial \boldsymbol{\xi}_{c}^{o}} & =
\frac{\partial \mathbf{T}_o^c(0\mathbin{:}2, 3)}{\partial 
	\boldsymbol{\xi}_{c}^{o}} \\ & =
\frac{\partial {\mathbf{T}_c^o}^{-1}(0\mathbin{:}2, 3)}{\partial 
	\boldsymbol{\xi}_{c}^{o}} \\ & =
\frac{\partial \big((exp(\hat{\delta 
		\boldsymbol{\xi}})\mathbf{T}_c^o)^{-1}(0\mathbin{:}2, 3)\big)}{\partial 
	(\delta \boldsymbol{\xi})} \\ & =
\frac{\partial \big((\mathbf{T}_o^c exp(-\hat{\delta 
		\boldsymbol{\xi}}))(0\mathbin{:}2, 3)\big)}{\partial (\delta 
	\boldsymbol{\xi})} \\ & =
\big(\mathbf{T}_o^c (-\frac{\partial exp(\hat{\delta 
		\boldsymbol{\xi}})}{\partial (\delta \boldsymbol{\xi})}) \big)(0\mathbin{:}2, 3) 
		\label{eq:generator1}\\ & 
=
-[(\mathbf{T}_o^c \mathbf{G}_0)(0\mathbin{:}2, 3), (\mathbf{T}_o^c 
\mathbf{G}_1)(0\mathbin{:}2, 3),..., (\mathbf{T}_o^c \mathbf{G}_5)(0\mathbin{:}2, 
3)],\label{eq:generator2}
\end{align}
where we use $(0\mathbin{:}2, 3)$ to denote the operation of getting the 
translation part from the corresponding matrix; $\mathbf{G}_0,...,\mathbf{G}_5$ 
are the infinitesimal generators of SE(3). Assuming the first three elements in the twist 
coordinate correspond to the translation part and the last three correspond to the rotation part, 
the infinitesimal generators are defined as
\begin{align}
G_0 = \begin{bmatrix}
\phantom{-}0 & \phantom{-}0 & \phantom{-}0 & \phantom{-}1\\
\phantom{-}0 & \phantom{-}0 & \phantom{-}0 & \phantom{-}0\\
\phantom{-}0 & \phantom{-}0 & \phantom{-}0 & \phantom{-}0\\
\phantom{-}0 & \phantom{-}0 & \phantom{-}0 & \phantom{-}0
\end{bmatrix},\\
G_1 = \begin{bmatrix}
\phantom{-}0 & \phantom{-}0 & \phantom{-}0 & \phantom{-}0\\
\phantom{-}0 & \phantom{-}0 & \phantom{-}0 & \phantom{-}1\\
\phantom{-}0 & \phantom{-}0 & \phantom{-}0 & \phantom{-}0\\
\phantom{-}0 & \phantom{-}0 & \phantom{-}0 & \phantom{-}0
\end{bmatrix},\\			
G_2 = \begin{bmatrix}
\phantom{-}0 & \phantom{-}0 & \phantom{-}0 & \phantom{-}0\\
\phantom{-}0 & \phantom{-}0 & \phantom{-}0 & \phantom{-}0\\
\phantom{-}0 & \phantom{-}0 & \phantom{-}0 & \phantom{-}1\\
\phantom{-}0 & \phantom{-}0 & \phantom{-}0 & \phantom{-}0
\end{bmatrix},\\
G_3 = \begin{bmatrix}
\phantom{-}0 & \phantom{-}0 & \phantom{-}0 & \phantom{-}0\\
\phantom{-}0 & \phantom{-}0 &           -1 & \phantom{-}0\\
\phantom{-}0 & \phantom{-}1 & \phantom{-}0 & \phantom{-}0\\
\phantom{-}0 & \phantom{-}0 & \phantom{-}0 & \phantom{-}0
\end{bmatrix},\\
G_4 = \begin{bmatrix}
\phantom{-}0 & \phantom{-}0 & \phantom{-}1 & \phantom{-}0\\
\phantom{-}0 & \phantom{-}0 & \phantom{-}0 & \phantom{-}0\\
          -1 & \phantom{-}0 & \phantom{-}0 & \phantom{-}0\\
\phantom{-}0 & \phantom{-}0 & \phantom{-}0 & \phantom{-}0
\end{bmatrix},\\
G_5 = \begin{bmatrix}
\phantom{-}0 &           -1 & \phantom{-}0 & \phantom{-}0\\
\phantom{-}1 & \phantom{-}0 & \phantom{-}0 & \phantom{-}0\\
\phantom{-}0 & \phantom{-}0 & \phantom{-}0 & \phantom{-}0\\
\phantom{-}0 & \phantom{-}0 & \phantom{-}0 & \phantom{-}0
\end{bmatrix}.		  	
\end{align}

Lastly, the Jacobian with respect to $\mathbf{z}$ is
\begin{align}
\frac{\partial r_{trans}}{\partial \mathbf{z}} = \mathbf{0}.
\end{align}

\subsubsection{Jacobian of Rotation Prior Residuals.} The rotation prior 
residual can be reformulated to
\begin{align}
r_{rot} & =
1 - (\mathbf{R}_{o}^{c}[0, -1, 0]^{\top})^{\top} \mathbf{n}_{g} \\ & =
1 - [0, -1, 0] {\mathbf{R}_o^c}^{\top}\mathbf{n}_g \\ & =
1 + [0, 1, 0] \mathbf{R}_c^o\mathbf{n}_g \\ & =
1 + \mathbf{r}_2\mathbf{n}_g,
\end{align}
where $\mathbf{r}_1$ is the second row of $\mathbf{R}_c^o$. Therefore, the 
Jacobian with respect to $\boldsymbol{\xi}_{c}^{o}$ is
\begin{align}
\frac{\partial r_{rot}}{\partial \boldsymbol{\xi}_{c}^{o}} & =
\frac{\partial \mathbf{r}_1}{\partial \boldsymbol{\xi}_{c}^{o}}\mathbf{n}_g \\ 
& =
\frac{\partial \mathbf{T}_c^o(1,0\mathbin{:}2)}{\partial 
	\boldsymbol{\xi}_{c}^{o}}\mathbf{n}_g \\ & =
(\frac{\partial exp(\hat{\delta \boldsymbol{\xi}})}{\partial (\delta 
	\boldsymbol{\xi})} \mathbf{T}_c^o)(1,0\mathbin{:}2) \mathbf{n}_g \\ & =
[(\mathbf{G}_0\mathbf{T}_c^o)(1,0\mathbin{:}2),...,(\mathbf{G}_5\mathbf{T}_c^o)
(1,0\mathbin{:}2)]\mathbf{n}_g,
\end{align}
where $(1,0\mathbin{:}2)$ denotes the operation of getting the part 
corresponding to the second row of the rotation matrix. Lastly, we have
\begin{align}
\frac{\partial r_{rot}}{\partial \mathbf{z}} = \mathbf{0}.
\end{align}

\section{More Qualitative Results}
\label{more_res}
\begin{figure*}
	\centering
	\includegraphics[width=0.465\textwidth]{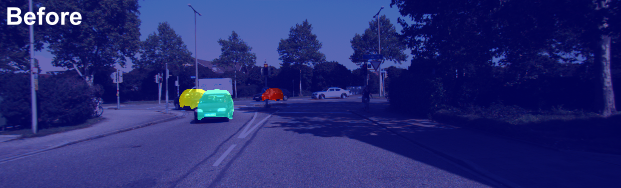}%
	\includegraphics[width=0.465\textwidth]{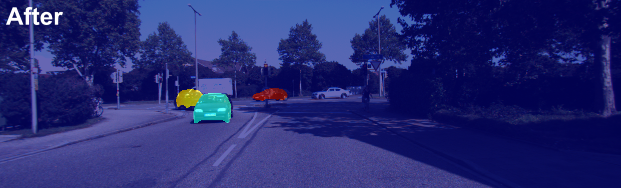}
	\includegraphics[width=0.186\textwidth]{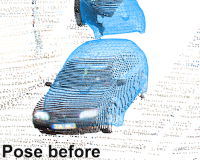}%
	\includegraphics[width=0.186\textwidth]{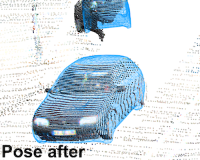}%
	\includegraphics[width=0.186\textwidth]{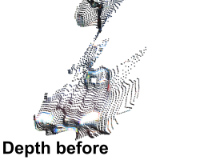}%
	\includegraphics[width=0.186\textwidth]{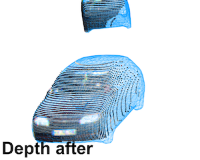}%
	\includegraphics[width=0.186\textwidth]{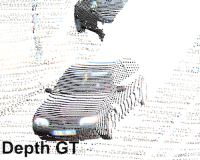}\vspace{1ex}
	\includegraphics[width=0.465\textwidth]{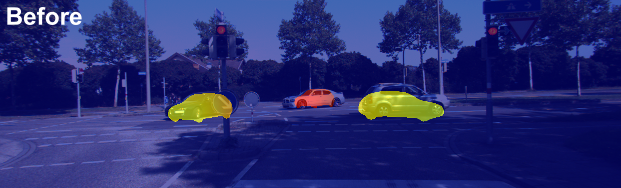}%
	\includegraphics[width=0.465\textwidth]{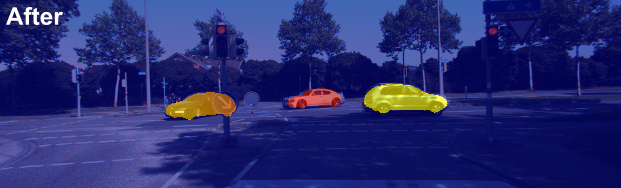}
	\includegraphics[width=0.186\textwidth]{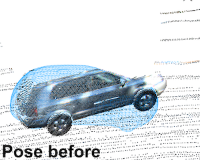}%
	\includegraphics[width=0.186\textwidth]{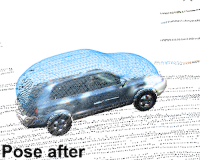}%
	\includegraphics[width=0.186\textwidth]{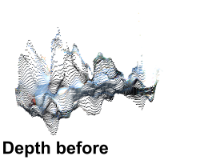}%
	\includegraphics[width=0.186\textwidth]{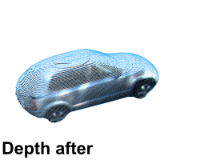}%
	\includegraphics[width=0.186\textwidth]{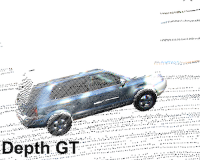}\vspace{1ex}
	\includegraphics[width=0.465\textwidth]{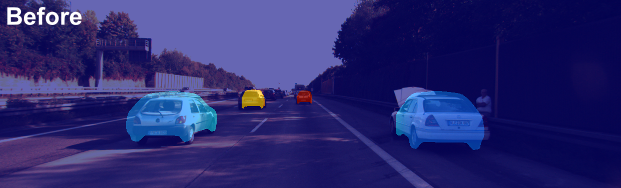}%
	\includegraphics[width=0.465\textwidth]{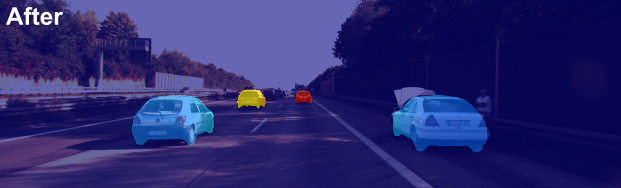}
	\includegraphics[width=0.186\textwidth]{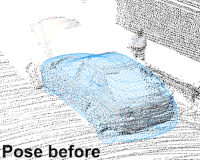}%
	\includegraphics[width=0.186\textwidth]{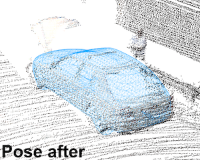}%
	\includegraphics[width=0.186\textwidth]{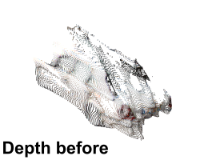}%
	\includegraphics[width=0.186\textwidth]{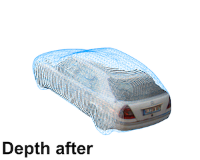}%
	\includegraphics[width=0.186\textwidth]{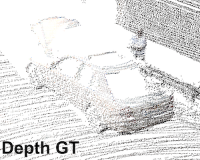}\vspace{1ex}
	\includegraphics[width=0.465\textwidth]{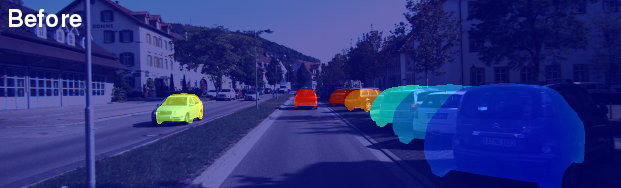}%
	\includegraphics[width=0.465\textwidth]{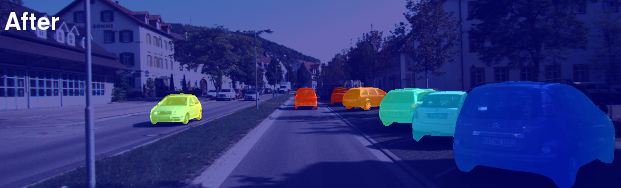}
	\includegraphics[width=0.186\textwidth]{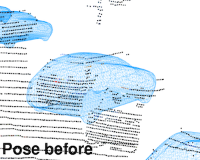}%
	\includegraphics[width=0.186\textwidth]{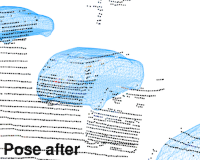}%
	\includegraphics[width=0.186\textwidth]{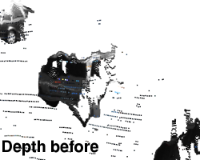}%
	\includegraphics[width=0.186\textwidth]{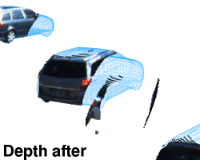}%
	\includegraphics[width=0.186\textwidth]{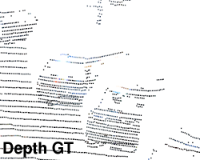}
	\caption{Qualitative results on 3D pose and shape refinements. Each two-row block shows the 
	results 
		on one stereo frame from the KITTI Stereo 2015 benchmark. Top row: The initial pose and 
		shape (left) and our results (right) projected to the left image. Bottom row: The 
		initial poses estimated by 3DOP and the poses refined by our method shown together with the 
		ground truth 3D point cloud (1st and 2nd); The following three images show the 3D points 
		estimated by ELAS (middle), our method (4th) and from the ground truth (last). Note that 
		the 3D point clouds are not used in our optimization. (better viewed electronically)}
	\label{fig1}
\end{figure*}

In Fig.~\ref{fig1} we qualitatively show the refinements on 3D pose and shape delivered by our 
method. The results on each stereo image pair are shown in each two-row block. In the first row we 
show the initial pose and shape estimates and our results projected onto the left image. In the 
second row, the initial pose (3DOP) and the estimated pose by our method are shown in the first two 
images, together with the ground truth 3D point cloud. In the following three images, the 3D point 
cloud estimated by ELAS and by our method, as well as the ground truth are shown, respectively.
As shown in the results, dense stereo matching results become extremely noisy on the strong 
non-lambertian reflective car surfaces. Our results avoid using such results for recovering the 3D 
poses and shapes of cars, instead it works directly on images by performing joint silhouette and 
photometric alignment. While it drastically improves the 3D shape reconstruction, it can also 
effectively recover the 3D poses of the objects. 

\begin{figure*}[]
	\centering
	\includegraphics[width=0.44\textwidth]{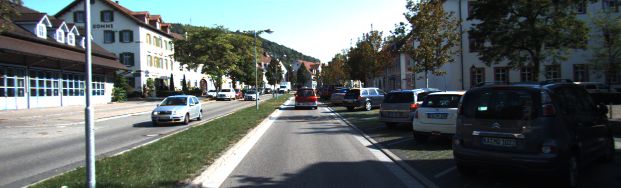}%
	\includegraphics[width=0.44\textwidth]{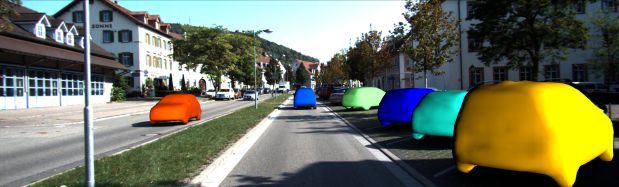}\vspace{1ex}
	
	\includegraphics[width=0.44\textwidth]{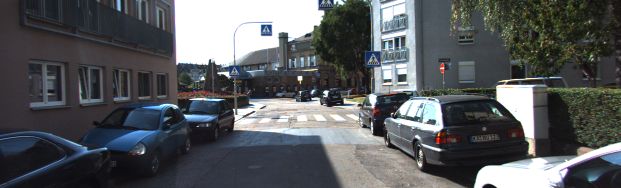}%
	\includegraphics[width=0.44\textwidth]{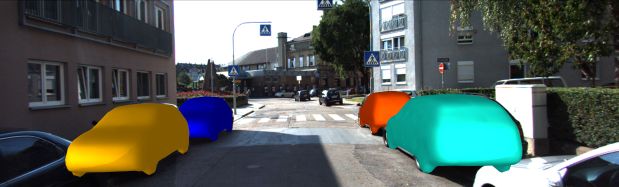}\vspace{1ex}
	
	\includegraphics[width=0.44\textwidth]{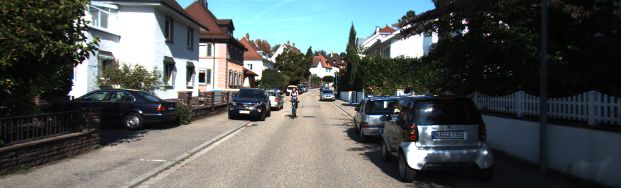}%
	\includegraphics[width=0.44\textwidth]{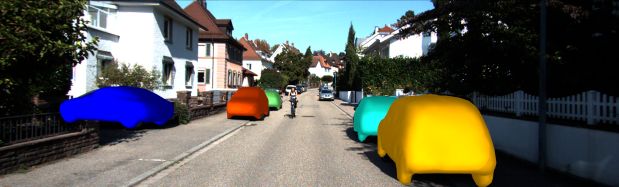}\vspace{1ex}
	
	\includegraphics[width=0.44\textwidth]{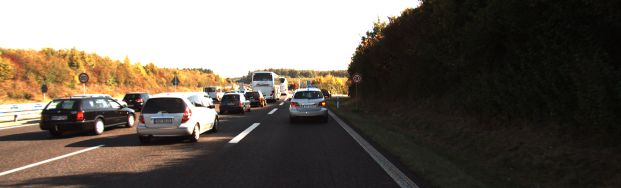}%
	\includegraphics[width=0.44\textwidth]{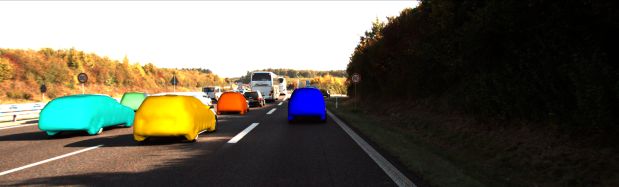}\vspace{1ex}
	
	\includegraphics[width=0.44\textwidth]{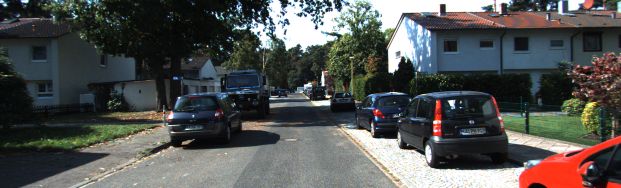}%
	\includegraphics[width=0.44\textwidth]{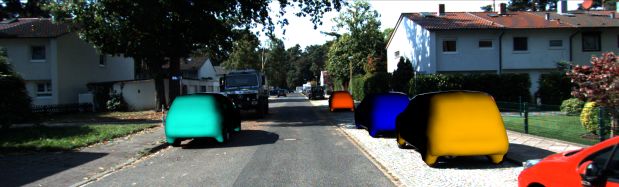}\vspace{1ex}
	
	\includegraphics[width=0.44\textwidth]{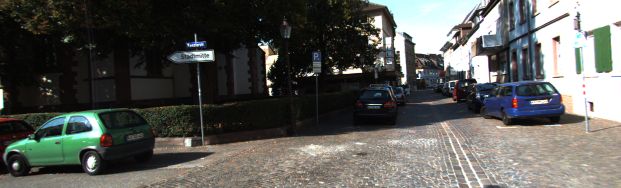}%
	\includegraphics[width=0.44\textwidth]{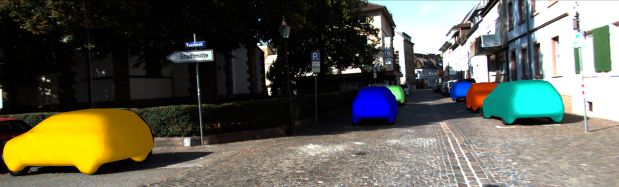}\vspace{1ex}
	
	\includegraphics[width=0.44\textwidth]{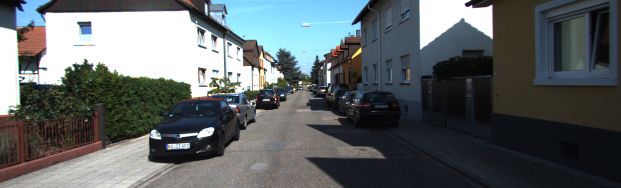}%
	\includegraphics[width=0.44\textwidth]{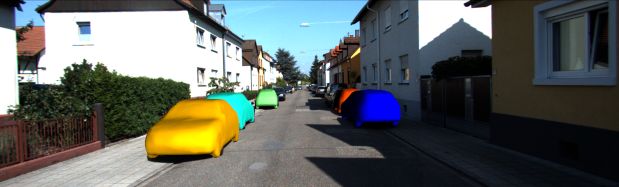}\vspace{1ex}
	
	\includegraphics[width=0.44\textwidth]{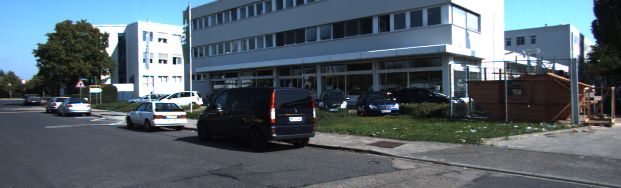}%
	\includegraphics[width=0.44\textwidth]{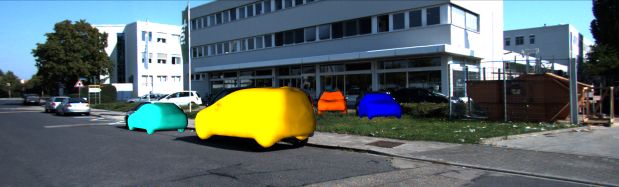}\vspace{1ex}
	
	\includegraphics[width=0.44\textwidth]{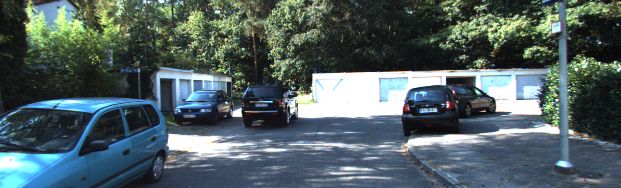}%
	\includegraphics[width=0.44\textwidth]{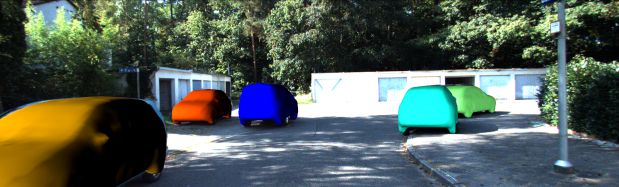}\vspace{1ex}
	
	\caption{Qualitative results on 3D pose and shape estimation.
	}\vspace{-1ex}
	\label{fig:eval_pose_quality1}
\end{figure*}

\begin{figure*}[]
	\centering
	\includegraphics[width=0.44\textwidth]{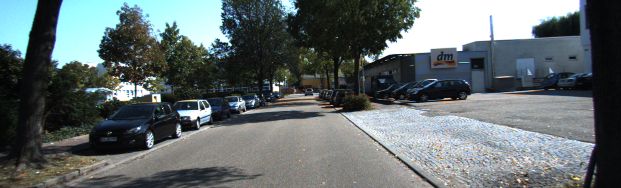}%
	\includegraphics[width=0.44\textwidth]{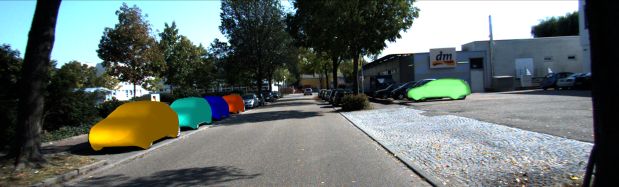}\vspace{1ex}
	
	\includegraphics[width=0.44\textwidth]{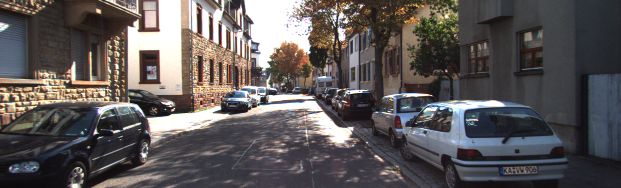}%
	\includegraphics[width=0.44\textwidth]{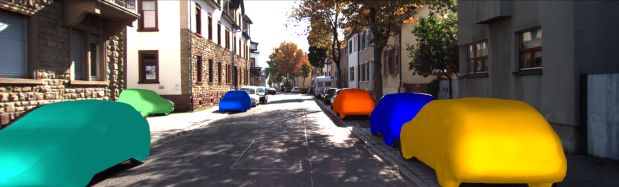}\vspace{1ex}
	
	\includegraphics[width=0.44\textwidth]{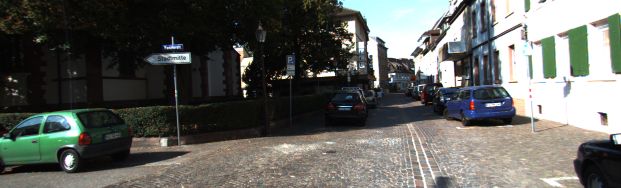}%
	\includegraphics[width=0.44\textwidth]{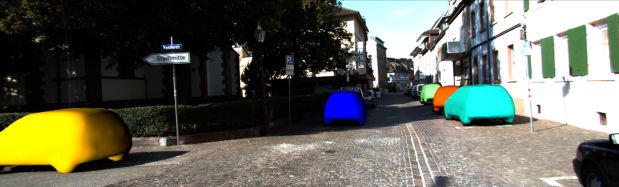}\vspace{1ex}	
	
	\includegraphics[width=0.44\textwidth]{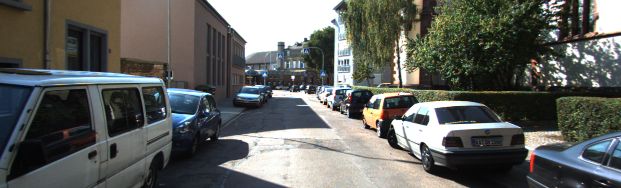}%
	\includegraphics[width=0.44\textwidth]{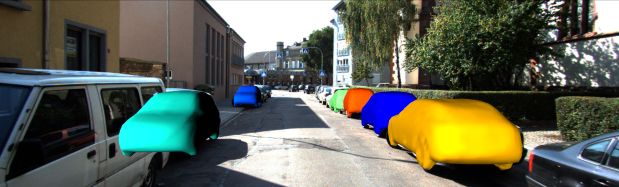}\vspace{1ex}
	
	\includegraphics[width=0.44\textwidth]{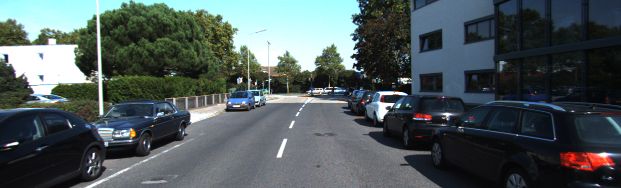}%
	\includegraphics[width=0.44\textwidth]{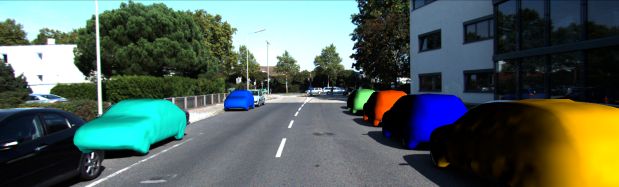}\vspace{1ex}
	
	\includegraphics[width=0.44\textwidth]{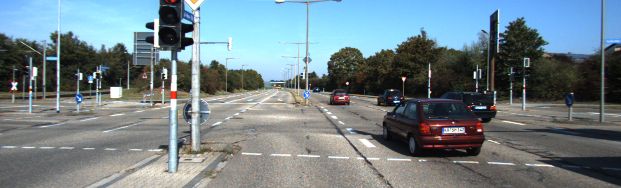}%
	\includegraphics[width=0.44\textwidth]{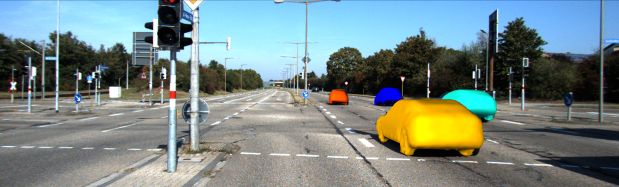}\vspace{1ex}
	
	\includegraphics[width=0.44\textwidth]{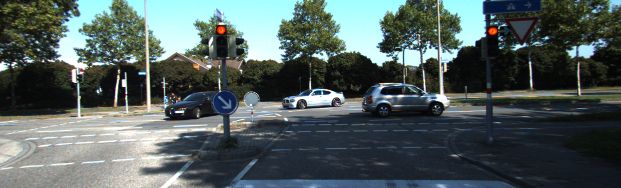}%
	\includegraphics[width=0.44\textwidth]{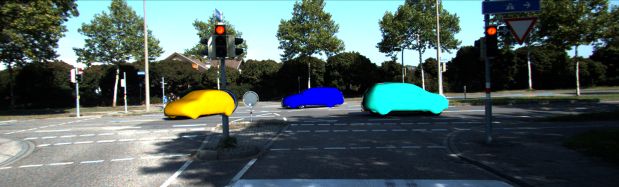}\vspace{1ex}
	
	\includegraphics[width=0.44\textwidth]{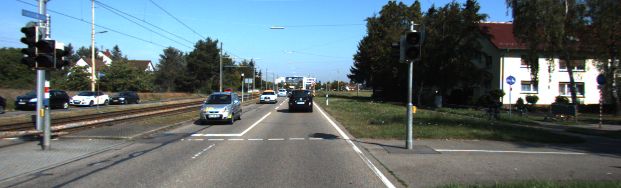}%
	\includegraphics[width=0.44\textwidth]{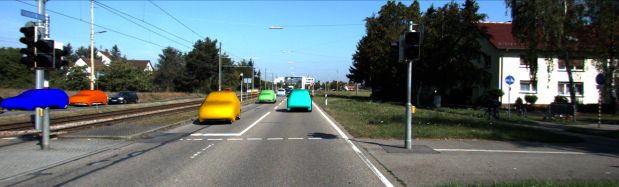}\vspace{1ex}
	
	\includegraphics[width=0.44\textwidth]{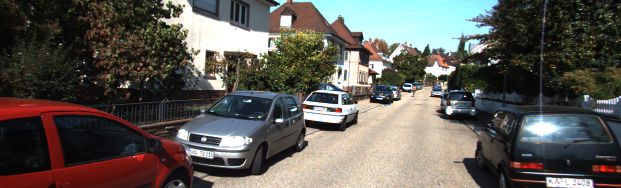}%
	\includegraphics[width=0.44\textwidth]{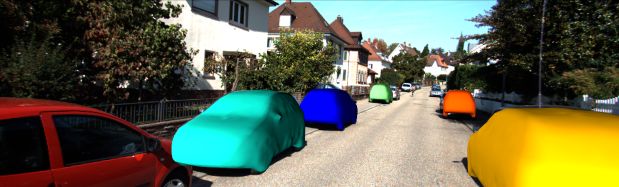}\vspace{1ex}
	
	\caption{Qualitative results on 3D pose and shape estimation (cont.).
	}\vspace{-1ex}
	\label{fig:eval_pose_quality2}
\end{figure*}

More qualitative results are listed in Fig.~\ref{fig:eval_pose_quality1} and 
\ref{fig:eval_pose_quality2}, where the input images are shown on the left and our results overlaid 
to the input images are shown on the right.

